\documentclass[10pt, letter, onecolumn]{arxiv}
\usepackage{amsmath,amssymb,amsfonts}%
\usepackage{amsthm}%
\usepackage{mathrsfs}%
\usepackage[title]{appendix}%
\usepackage{xcolor}%
\usepackage{textcomp}%
\usepackage{manyfoot}%
\usepackage{booktabs}%
\usepackage{algorithm}%
\usepackage{algpseudocode}
\usepackage{algorithmicx}%
\usepackage{listings}%
\usepackage{pifont}
\usepackage{utfsym}

\usepackage{makecell}
\usepackage{kantlipsum, lipsum}
\usepackage{dm-colors}
\usepackage{pstricks, pst-node}
\usepackage{verbatim}
\usepackage{multirow}
\usepackage{scalerel}
\usepackage{booktabs}
\usepackage{enumitem}
\usepackage{xspace}
\usepackage{bm}
\usepackage{bbm}
\usepackage{mathtools}
\usepackage{soul}
\usepackage{epsfig}
\usepackage{graphicx}
\usepackage{csquotes}
\usepackage{setspace}
\usepackage{colortbl}
\usepackage{threeparttable}
\usepackage{tabularx,ragged2e} 
\usepackage{placeins}
\usepackage[hang,flushmargin,symbol]{footmisc}
\usepackage[normalem]{ulem}
\useunder{\uline}{\ul}{}

\usepackage{varioref}
\usepackage[pagebackref=false,breaklinks=false,%
            colorlinks=true,bookmarks=true,citecolor=ourdarkblue,%
            urlcolor=ourdarkblue,linkcolor=ourdarkblue]{hyperref}
\usepackage[noabbrev,capitalize]{cleveref}
\usepackage{etoc}
\usepackage{lineno}
\usepackage{mdframed}
\usepackage{amsmath}

\DeclareMathOperator*{\argmax}{arg\,max}

\definecolor{mygray}{rgb}{0.85, 0.85, 0.85}

\title{\hspace{3pt}\Large{Boosting Pathology Foundation Models via Few-shot Prompt-tuning for Rare Cancer Subtyping}}

\author[1$\ast$]{Dexuan He}
\author[2$\ast$]{Xiao Zhou}
\author[3$\ast$]{Wenbin Guan}
\author[1]{Liyuan Zhang}
\author[4]{Xiaoman Zhang}
\author[1]{Sinuo Xu}
\author[5]{\\ \vspace{0.12cm} Ge Wang}
\author[3]{Lifeng Wang}
\author[6]{Xiaojun Yuan}
\author[7]{Xin Sun}
\author[1,2]{Yanfeng Wang}
\author[8 $^\dagger$]{\\ \vspace{0.12cm} Kun Sun}
\author[1,2 $^\dagger$]{Ya Zhang}
\author[1,2 $^\dagger$]{Weidi Xie}

\affil[1]{\normalsize School of Artificial Intelligence, Shanghai Jiao Tong University \par}
\affil[2]{\normalsize Shanghai Artificial Intelligence Laboratory \par}
\affil[3]{\normalsize Department of Pathology, Xinhua Hospital \par Affiliated to Shanghai Jiao Tong University School of Medicine \par}
\affil[4]{\normalsize Department of Biomedical Informatics, Harvard Medical School \par}
\affil[5]{\normalsize Department of Oral Pathology, Shanghai Ninth People's Hospital, \par Shanghai Jiao Tong University School of Medicine \par}
\affil[6]{\normalsize Department of Pediatric Hematology/Oncology, Xinhua Hospital \par Affiliated to Shanghai Jiao Tong University School of Medicine \par}
\affil[7]{\normalsize Clinical Research and Innovation Unit, Xinhua Hospital \par Affiliated to Shanghai Jiao Tong University School of Medicine \par}
\affil[8]{\normalsize Department of Pediatric Cardiology, Xinhua Hospital \par Affiliated to Shanghai Jiao Tong University School of Medicine \par}

\affil[$\ast$]{\normalsize Equal contributions \hspace{1cm}}
\affil[$\dagger$]{\normalsize Corresponding authors\authorcr Kun Sun: sunkun@xinhuamed.com.cn; \authorcr Ya Zhang: ya\_zhang@sjtu.edu.cn; \hspace{0.5cm} Weidi Xie: weidi@sjtu.edu.cn}

\begin{document}

\nolinenumbers 

\begin{abstract}

Rare cancers comprise 20–25\% of all malignancies but face major diagnostic challenges due to limited expert availability—especially in pediatric oncology, 
where they represent over 70\% of cases. 
While pathology vision-language~(VL) foundation models show promising zero-shot capabilities for common cancer subtyping, their clinical performance for rare cancers remains limited.
Existing multi-instance learning (MIL) methods rely only on visual features, overlooking cross-modal knowledge and compromising interpretability critical for rare cancer diagnosis.
To address this limitation, we propose \textbf{PathPT}, a novel framework that fully exploits the potential of vision-language pathology foundation models through spatially-aware visual aggregation and task-specific prompt tuning. 
Unlike conventional MIL, PathPT converts WSI-level supervision into fine-grained tile-level guidance by leveraging the zero-shot capabilities of VL models, thereby preserving localization on cancerous regions and enabling cross-modal reasoning through prompts aligned with histopathological semantics. 
We benchmark PathPT on eight rare cancer datasets (four adult and four pediatric) spanning 56 subtypes and 2,910 WSIs, as well as three common cancer datasets, evaluating four state-of-the-art VL models and four MIL frameworks under three few-shot settings. 
Results show that PathPT consistently delivers superior performance, achieving substantial gains in subtyping accuracy and cancerous region grounding ability.
This work advances AI-assisted diagnosis for rare cancers, offering a scalable solution for improving subtyping accuracy in settings with limited access to specialized expertise.
\end{abstract}

\maketitle
\section{Introduction}

Rare cancers, while individually uncommon, collectively account for 20–25\% of all malignancies and pose significant diagnostic challenges due to a lack of clinical expertise and limited reference cases~\cite{desantis2017burden}. These challenges are particularly pronounced in pediatric oncology, where rare tumors comprise over 70\% of diagnoses~\cite{desantis2017burden}, and timely, accurate subtyping is essential for effective treatment~\cite{butler2021recent, ni2022socioeconomic}. 
In many settings, for example, underserved regions and specialized clinics, the scarcity of experienced pathologists further underscores the urgent need for automated diagnostic tools capable of reliable performance under conditions of data scarcity and minimal supervision.

Recent advances in deep learning, particularly in self-supervised~\cite{chen2020improved, he2022mae, zhou2021ibot, caron2021dino} and vision-language (VL)~\cite{radford2021clip, zhai2023siglip, yu2022coca} pretraining, have driven the development of pathology foundation models~\cite{ chen2024uni, huang2023visual, ikezogwo2024quilt, ma2024towards, nechaev2024hibou, shaikovski2024prism, sun2025cpath, wang2022transformer, xu2024whole,  xiang2025vision,  xu2024multimodal, yang2025foundation, zhou2024kep}. 
Vision-only models such as Virchow~\cite{vorontsov2024virchow} show promise for rare cancer detection through fine-tuning, while vision-language models like TITAN~\cite{ding2024multimodal}, CONCH~\cite{lu2024visual}, and KEEP~\cite{zhou2024keep} leverage multimodal knowledge for zero-shot classification. However, despite success on benchmark datasets, these models face significant barriers to deployment in real-world settings for rare cancers, where annotated data and clinical expertise are scarce, zero-shot performance remains insufficient, and existing methods struggle to generalize.

The predominant paradigm for adapting vision-language foundation models to rare cancer subtyping is multi-instance learning (MIL)~\cite{Ilse2018abmil, lu2023visual, lu2021data, qu2024rise, shao2021transmil, shi2024vila, zhu2024dgr}, which aggregate tile-level visual features extracted from whole slide images (WSIs) and train classifiers under slide-level supervision.
While effective, this approach has two critical limitations.
\textbf{First}, MIL relies exclusively on the visual encoder, neglecting the semantic and cross-modal reasoning capabilities of the textual encoder. 
\textbf{Second}, slide-level supervision provides limited spatial guidance, hindering the ability to capture fine-grained, region-specific patterns essential for rare cancer subtyping. These limitations highlight the need for frameworks that can fully harness the prior knowledge in vision-language models while addressing the challenges of weak supervision.

To address these limitations, we propose \textbf{PathPT}—a novel framework that aims to fully harnesses the potential of pre-trained vision-language models for rare cancer subtyping under few-shot learning conditions. PathPT introduces three core innovations: \textbf{(i) spatially-aware visual aggregation}: 
it employs a lightweight aggregator that explicitly models short- and long-range dependencies across tissue regions, capturing complex morphological patterns critical for rare subtype diagnosis;
\textbf{(ii) task-adaptive prompt tuning}: PathPT replaces static, handcrafted language prompts with learnable textual tokens, optimized end-to-end to align with histopathological semantics, thereby preserving the prior knowledge of existing vision-language models;
\textbf{(iii) tile-level supervision from slide-level labels}: leveraging the zero-shot grounding ability of vision-language foundation models, PathPT transforms weak slide-level annotations into fine-grained tile-level pseudo-labels. This enables precise spatial learning, significantly improving both classification accuracy and cancerous region grounding performance.

\begin{figure}[!t]
  \centering
  \includegraphics[width=\textwidth]{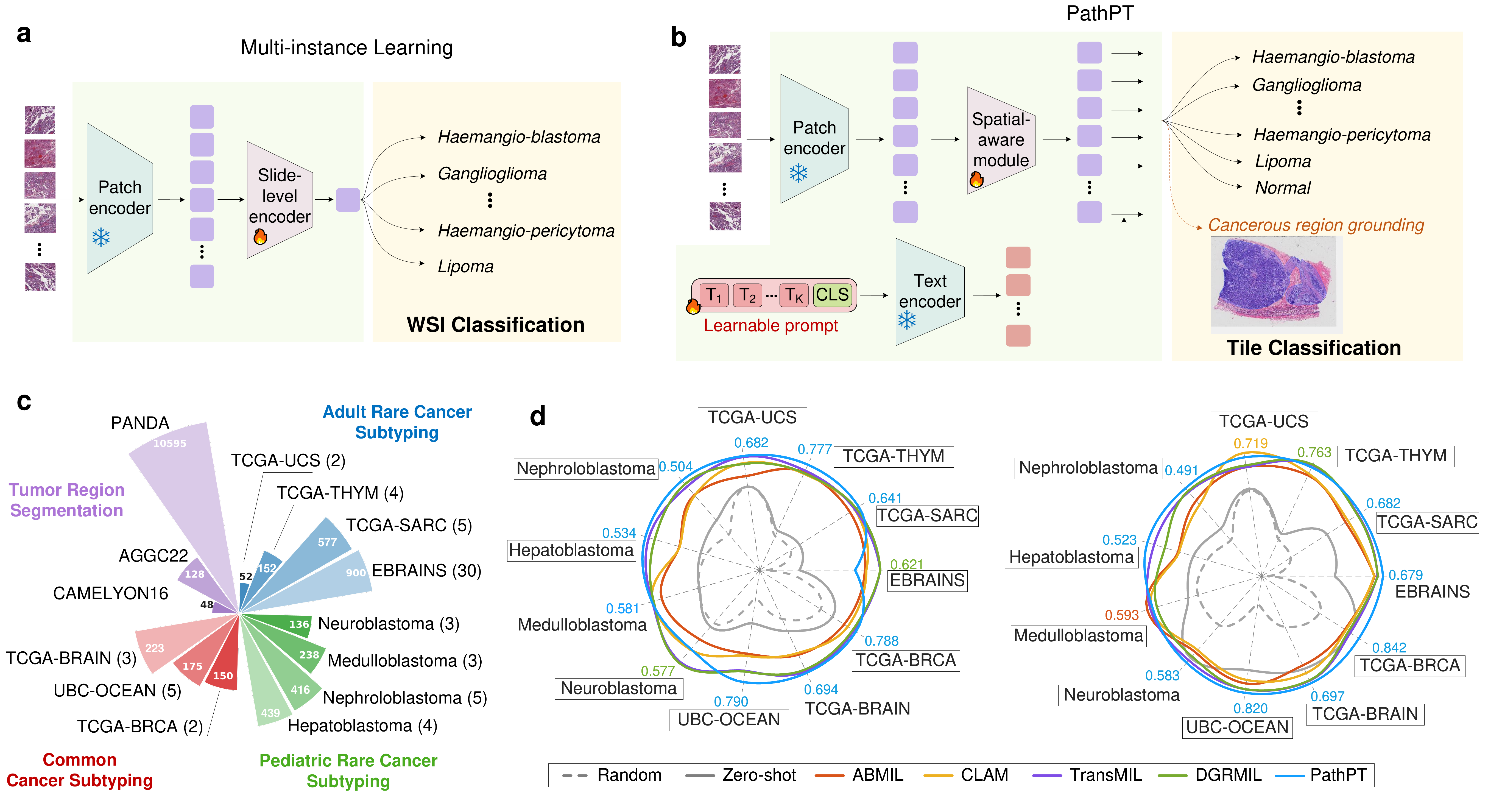}
  \caption{\textbf{Overview of this study}. 
  \textbf{a}. The pipeline of multi-instance learning framework, where tile-level visual features are aggregated into a slide-level representation to perform WSI classification. 
  \textbf{b}. The pipeline of PathPT, where both visual and textual features are employed to achieve fine-grained tile classification.
  \textbf{c}. The statistics of datasets utilized in this study, where the numbers in parentheses indicate the count of cancer subtypes, and the numbers within each sector represent the quantity of WSIs.
  \textbf{d}. The average performance of different 10-shot learning approaches across all subtyping tasks using CONCH (left) and KEEP (right) as the base models. The values denote the highest performance, and the colors indicate the corresponding best-performing methods. PathPT-CONCH and PathPT-KEEP achieve the best performance on 9/11 and 8/11 cancer subtyping benchmarks, respectively. Additional figures of performance comparison can be found in Supplementary Figure~\ref{fig:teaser_supp}.}
  \label{fig:teaser}
\end{figure}

For evaluation, we establish \textbf{eight rare cancer benchmarks}—four adult and four pediatric—spanning 56 subtypes and 2,920 WSIs, as well as three common cancer benchmarks~(Figure~\ref{fig:teaser}c).
We first benchmark four widely-used MIL frameworks—ABMIL~\cite{Ilse2018abmil}, CLAM~\cite{lu2021data}, TransMIL~\cite{shao2021transmil}, and DGRMIL~\cite{zhu2024dgr}—using tile-level features extracted from state-of-the-art vision-language (VL) models (PLIP~\cite{huang2023visual}, CONCH~\cite{lu2024visual}, MUSK~\cite{xiang2025vision}, and KEEP~\cite{zhou2024keep}). These frameworks operate as vision-only models, that trains multi-class classification on the 
aggregated tile-level visual features with slide-level supervision. This provides a baseline to assess the limitations of MIL in leveraging the prior knowledge of vision-language models for few-shot learning in rare cancer subtyping. We then evaluate PathPT under the same experimental conditions to directly compare its performance against the MIL baselines. 

While all methods improve over the zero-shot baselines, 
PathPT consistently delivers superior performance~(Figure~\ref{fig:teaser}d), 
achieving substantial gains in accuracy and interpretability. 
Notably, with KEEP~\cite{zhou2024keep} as the backbone, PathPT achieves 0.679 balanced accuracy on the EBRAINS~\cite{roetzer2022ebrains} dataset (30 subtypes, 10-shot), outperforming all MIL baselines. Further validation is conducted using three common cancer datasets (10 subtypes, 548 WSIs) under few-shot conditions, where PathPT continues to demonstrate strong generalizability and performance gains. Additionally, PathPT achieves significant improvements in tumor region segmentation, even in the challenging 1-shot setting, confirming its ability to leverage minimal supervision for precise spatial localization.

\section{Results}

\subsection{Overview on Benchmarking Approaches}

To comprehensively evaluate the performance of different computational methods on rare cancer subtype identification, particularly in few-shot learning regimes, we investigate two main lines of approaches: established multi-instance learning (MIL) frameworks and our proposed PathPT. 
In both sets of methods, we adopt the pre-extracted, frozen tile-level vision features derived from whole slide images (WSIs). We employed four distinct tile-level vision-language foundation models. Specifically, PLIP~\cite{huang2023visual} was trained using 200K image-text pairs from Twitter via vision-language contrastive learning. The other three models, {\em e.g.}, CONCH~\cite{lu2024visual}, MUSK~\cite{xiang2025vision}, and KEEP~\cite{zhou2024keep} all initialized their vision encoders with self-supervised weights before vision-language alignment training. CONCH utilized a {\em private} dataset of 1M image-text pairs, while MUSK and KEEP leveraged the {\em public} Quilt1M~\cite{ikezogwo2024quilt} dataset (1M image-text pairs). Notably, KEEP further incorporated disease knowledge injection to enhance performance.

These extracted and frozen features serve as input for four representative multi-instance learning (MIL) frameworks~(Figure~\ref{fig:teaser}a). ABMIL~\cite{Ilse2018abmil} employs attention mechanisms to weight and aggregate patch features for whole-slide classification. CLAM~\cite{lu2021data} integrates clustering constraints with attention to group patches, thereby enhancing feature representation. TransMIL~\cite{shao2021transmil} leverages self-attention with positional encoding to capture long-range dependencies among patches. DGRMIL~\cite{zhu2024dgr} introduces learnable global vectors and cross-attention to model instance diversity, thereby capturing tissue variability in WSI classification.

In contrast, our proposed framework, PathPT (Figure~\ref{fig:teaser}b), jointly exploits the inherent prior knowledge within pre-trained vision-language foundation models. Specifically, it introduces a spatial-aware module to capture both local and global tile contexts, and replaces hand-crafted prompts with learnable vectors optimized end-to-end with the frozen text encoder of a vision–language model. To enable tile-level training, PathPT leverages pre-trained vision-language pathology foundation models to perform zero-shot predictions on individual tiles within each WSI, then uses tiles whose predictions are normal or align with the WSI-level label for fine-grained training. This unique capability allows PathPT to distinguish cancerous from normal tissue regions, thereby localizing tumor regions across different cancer subtypes within WSIs and providing interpretable spatial localization for pathological diagnosis.

We subsequently compared PathPT against these established MIL baselines across multiple datasets, encompassing both rare and common cancer cohorts. These comprehensive evaluations demonstrate PathPT’s superior ability to operate effectively in few-shot learning regimes, robustly leverage prior knowledge from pre-trained vision-language models, and significantly improve diagnostic performance across both classification and segmentation tasks.

\subsection{Benchmark on Rare Adult Cancer Subtyping}

We evaluated PathPT and established multi-instance learning (MIL) methods on rare tumor cohorts from EBRAINS~\cite{roetzer2022ebrains} and TCGA datasets. Detailed information on these two datasets can be found in Supplementary Table~\ref{tab:data_ebrains} and~\ref{tab:data_adult}.
Following prior protocols~({\em e.g.}, CONCH, KEEP), we selected 30 subtypes with at least 30 whole-slide images (WSIs) each. For every subtype, 15 WSIs were randomly assigned to the training set, with the remainder reserved for testing. 

To assess \textbf{few-shot performance}, we sampled 1, 5, or 10 WSIs per subtype from the training set and repeated each experiment 10 times to account for variance. PathPT was applied on four vision–language foundations (PLIP, MUSK, CONCH, KEEP) and compared against four MIL baselines: ABMIL~\cite{Ilse2018abmil}, CLAM~\cite{lu2021data}, TransMIL~\cite{shao2021transmil}, and DGRMIL~\cite{zhu2024dgr}.

\begin{figure}[!t]
  \centering
  \includegraphics[width=\textwidth]{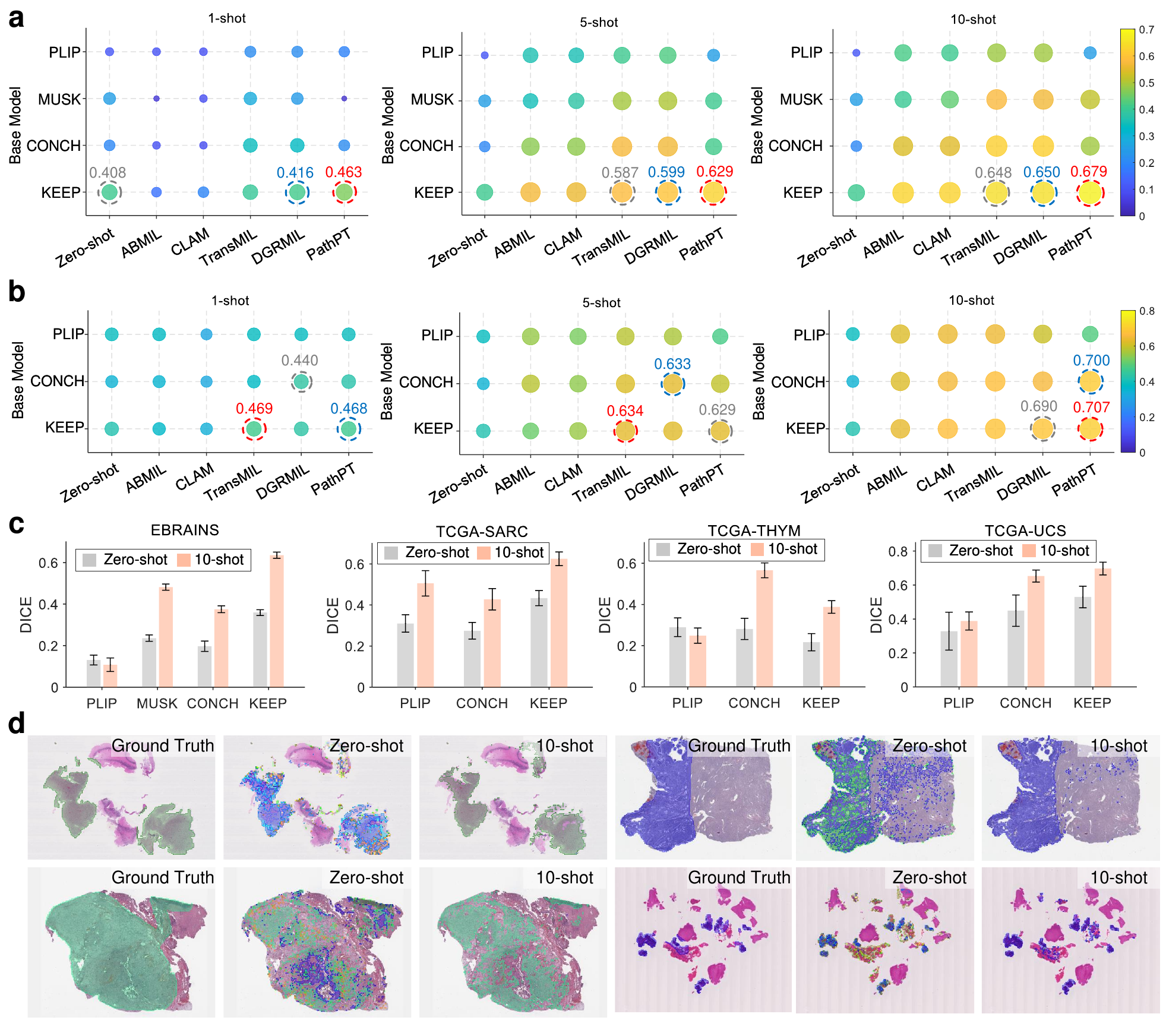}
  \caption{\textbf{Rare adult cancer subtyping results}. 
  \textbf{a}. The performance comparison on EBRAINS (brain tumors, 30 subtypes) between PathPT and MIL baselines, including ABMIL, CLAM, TransMIL, and DGRMIL, using different base models (PLIP, MUSK, CONCH, and KEEP), in 1-shot, 5-shot, and 10-shot settings. Red, blue, and gray circles highlight the best, second-best, and third-best performance, respectively, across all base models and few-shot learning methods.
  \textbf{b}. Average performance on rare adult cancer benchmark datasets from TCGA, including TCGA-SARC (sarcoma; 5 subtypes), TCGA-THYM (thymoma; 4 subtypes), and TCGA-UCS (uterine carcinosarcoma; 2 subtypes). MUSK is excluded because it was pretrained on TCGA. Red and blue circles highlight the best, second-best, and third-best performance respectively, across all base models and few-shot learning methods.
  \textbf{c}. The DICE score comparison between zero-shot and PathPT fine-tuning in cancerous region segmentation on adult rare cancers.
  \textbf{d}. The segmentation visualization of exemplary WSI. Different colors of masks suggest different subtypes. More segmentation visualizations can be found in Supplementary Figure~\ref{fig:app_visual_sub_adult3},~\ref{fig:app_visual_sub_adult2} and~\ref{fig:app_visual_sub_adult1}.}
  \label{fig:result_rare_adult_sub}
\end{figure}

\textbf{Results on EBRAINS.}

As shown in Figure~\ref{fig:result_rare_adult_sub}a, zero-shot classification performance was limited across all foundation models, 
with average balanced accuracy between 0.1 and 0.4. Among MIL methods, TransMIL and DGRMIL consistently improved accuracy, particularly when leveraging KEEP-derived features. Similarly, PathPT with KEEP (PathPT–KEEP) surpassed all MIL baselines and other PathPT variants, achieving a median balanced accuracy of 0.679 in the 10-shot setting—a 0.271 absolute gain over its zero-shot baseline. Detailed results on performance can be found in Supplementary Figure~\ref{fig:result_rare_adult_supp} and Supplementary Table~\ref{tab:ebrains}.

In contrast, PathPT based on PLIP, MUSK, or CONCH underperformed MIL baselines, likely due to the relatively poor performance of the original vision-language model, {\em i.e.}, tile-level zero-shot predictions. 
To validate this assumption, we invite pathologists to annotate tumor regions for each WSI in the few-shot training set. 
The resulting DICE scores of zero-shot tumor grounding~(Figure~\ref{fig:result_rare_adult_sub}c) confirmed that KEEP provided the most reliable grounding. Representative segmentation visualizations are shown in Figure~\ref{fig:result_rare_adult_sub}d and Supplementary Figure~\ref{fig:app_visual_sub_adult3}.

\textbf{Results on Rare Cancers from TCGA.}

We further benchmarked on rare cancers from TCGA\footnote{https://portal.gdc.cancer.gov/} dataset: 
TCGA-SARC (sarcoma), TCGA-THYM (thymoma), and TCGA-UCS (uterine carcinoma). 
Sarcoma comprises \textbf{five subtypes}: dedifferentiated liposarcoma, leiomyosarcoma, malignant peripheral nerve sheath tumor, myxofibrosarcoma, and undifferentiated pleomorphic sarcoma. Thymoma includes \textbf{four subtypes}: type A, type AB, type B1, and type B2. 
Uterine carcinoma has \textbf{two subtypes}: heterologous and homologous. Average performance across these datasets is shown in Figure~\ref{fig:result_rare_adult_sub}b; subtype-specific distributions appear in Supplementary Figure~\ref{fig:result_rare_adult_supp}; detialed information on these datasets can be found in Supplementary Table~\ref{tab:data_adult}.

It can be seen that relative to the zero-shot baselines, both 5-shot and 10-shot fine-tuning yield consistent performance gains across all base models and adaptatoion methods, with additional improvements from 5 to 10 shots. 
Within this overall trend, KEEP emerges as the strongest base encoder, delivering the best results for both the MIL baselines and PathPT. Building on stronger backbones, PathPT is competitive in the extreme low-data regime, performing on par with TransMIL at 1 shot, and becomes the leading approach when more supervision is available, achieving the top result with KEEP and the second-best with CONCH under the 10-shot setting. 
The notable exception occurs with PLIP, with the effectiveness hinged by the performance of original vision-language model, as shown in Figure~\ref{fig:result_rare_adult_sub}c, PLIP exhibits limited zero-shot grounding on TCGA-SARC and TCGA-UCS, leading to low-quality pseudo labels that cap PathPT’s gains.

\begin{figure}[!t]
  \centering
  \includegraphics[width=\textwidth]{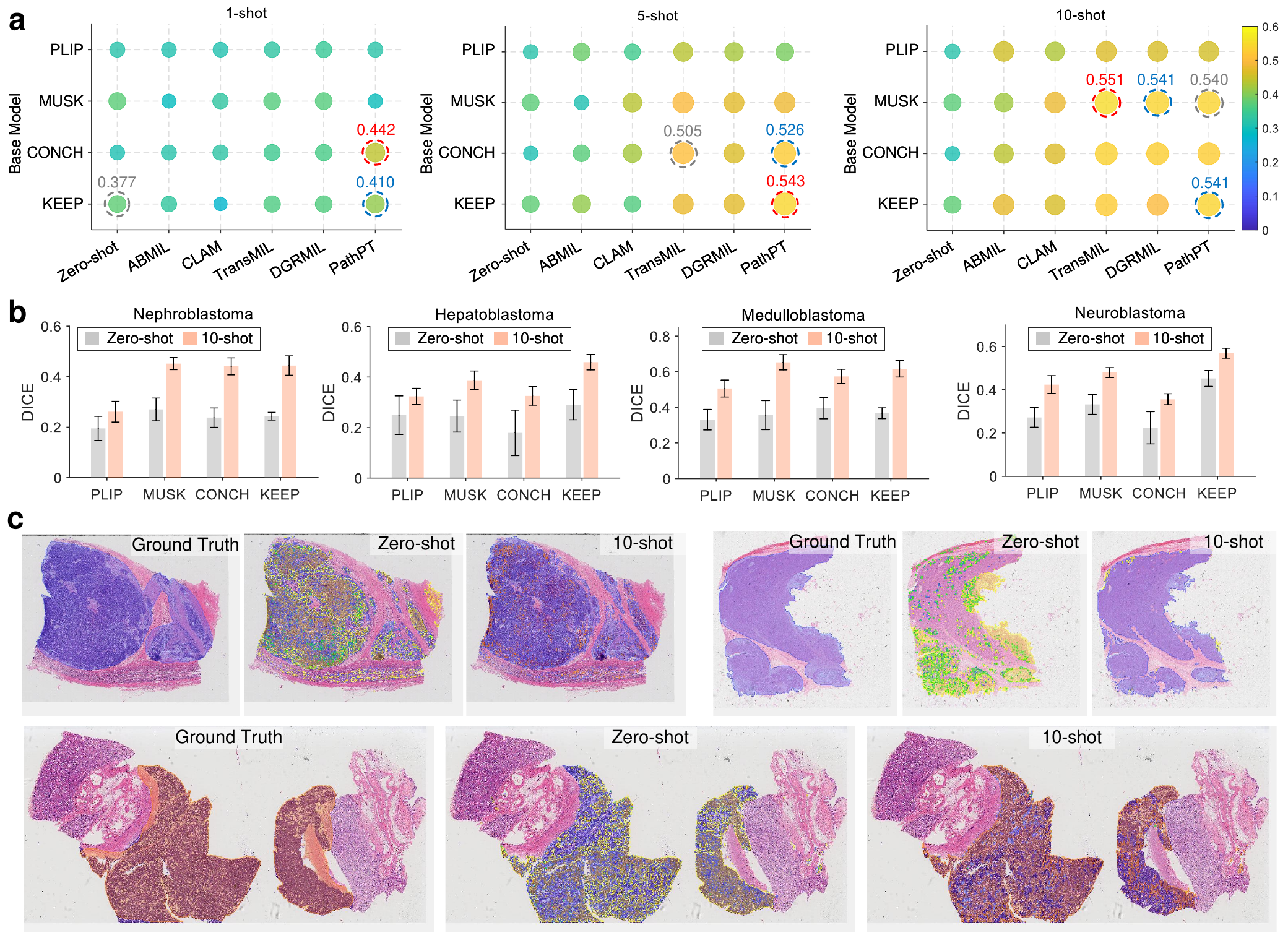}
  \caption{\textbf{Rare pediatric cancer subtyping results}. 
  \textbf{a}. Average performance on pediatric rare cancer benchmark datasets, including nephroblastoma (5 subtypes), hepatoblastoma(4 subtypes), medulloblastoma (3 subtypes), and neuroblastoma (3 subtypes). Red, blue, and gray circles highlight the best, second-best, and third-best performance, respectively, across all base models and few-shot learning methods.
  \textbf{b}. The DICE score comparison between zero-shot and PathPT fine-tuning in cancerous region segmentation on different pediatric cancers.
  \textbf{c}. The segmentation visualization of exemplary whole slide images. Different colors of masks suggest different subtypes. More segmentation visualizations can be found in Supplementary Figure~\ref{fig:app_visual_sub_child1} and~\ref{fig:app_visual_sub_child2}.}
  \label{fig:result_rare_pediatric_sub}
\end{figure}

\subsection{Benchmark on Rare Pediatric Cancer Subtyping}

To evaluate performance across pediatric cohorts, we curated 1,232 WSIs from patients at Xinhua Hospital, Shanghai Jiao Tong University, 
School of Medicine. The dataset spans four major cancer types and their subtypes: 
\textbf{(i) Nephroblastoma} (416 WSIs): blastemal (12); stromal (90); mixed blastemal–epithelial (57); mixed blastemal–stromal (53); mixed blastemal–epithelial–stromal (204); 
\textbf{(ii) Hepatoblastoma} (439 WSIs): epithelial macrotrabecular (10); mixed epithelial and mesenchymal (76); epithelial mixed fetal and embryonal (175); pure fetal with low mitotic activity (178); \textbf{(ii) Medulloblastoma} (238 WSIs): large cell/anaplastic (11); desmoplastic nodular (30); classic medulloblastoma~(197);
\textbf{(iv) Neuroblastoma} (136 WSIs): ganglioneuroblastoma, intermixed (32 cases); differentiating neuroblastoma (35); poorly differentiated neuroblastoma (69). Statistics of these datasets can be found in Supplementary Table~\ref{tab:data_child}.

Following the adult rare-cancer protocol, we sampled 15 WSIs per subtype for training and held out the remaining for testing. For subtypes with fewer than 15 WSIs, we split cases approximately 1:1. Few-shot performance was assessed with 1, 5, or 10 WSIs per subtype, each repeated 10 times with different samples. For subtypes with fewer than $k$ eligible training WSIs, we sampled with replacement to satisfy the $k$-shot setting.

\textbf{Results on Rare Pediatric Cancers.}

The average performance across the four cancer types is shown in Figure~\ref{fig:result_rare_pediatric_sub}a, with more breakdown details in Supplementary Figure~\ref{fig:result_rare_children_supp} and Supplementary Table~\ref{tab:nephro}-~\ref{tab:neuro}. Across model–method pairs, both 5-shot and 10-shot fine-tuning consistently outperformed zero-shot baselines, with a larger gain from 1 to 5 shots than from 5 to 10 (global best/runner-up balanced accuracy: 0.442/0.410 at 1-shot, 0.543/0.528 at 5-shot, 0.551/0.541 at 10-shot). 
The muted improvement from 5 to 10 shots can be attributed to several subtypes having at most five unique training WSIs; consequently, the 10-shot condition entails sampling with replacement, which does not introduce additional unique training information and therefore yields limited incremental benefit.

At the method level, PathPT surpassed all MIL baselines in the 1-shot and 5-shot settings. By 10 shots, PathPT, TransMIL, and DGRMIL converged to similar performance (0.54–0.55), suggesting a ceiling imposed by limited data. At the base-model level, KEEP and CONCH paired robustly with multiple methods, whereas MUSK achieved the overall best result with DGRMIL at 10-shot (0.551). In contrast, PLIP consistently lagged, due to the poor performance of the original vision-lanaguage model, yielding less discriminative features and limiting downstream adaptation.

To further assess the grounding ability, we also invited experienced pathologists to annotate tumor region with masks for each WSI in each of the training set. As shown in Figure~\ref{fig:result_rare_pediatric_sub}b, 
PathPT enables to effectively identify cancerous tiles with the supervision of WSI-level labels, resulting in superior grounding performance across all cancer types. Representative visualizations are presented in Figure~\ref{fig:result_rare_pediatric_sub}c, and Supplementary Figure~\ref{fig:app_visual_sub_child1},~\ref{fig:app_visual_sub_child2}. These results underscore PathPT’s robustness in spatial grounding with different base models.

\begin{figure}[!t]
  \centering
  \includegraphics[width=\textwidth]{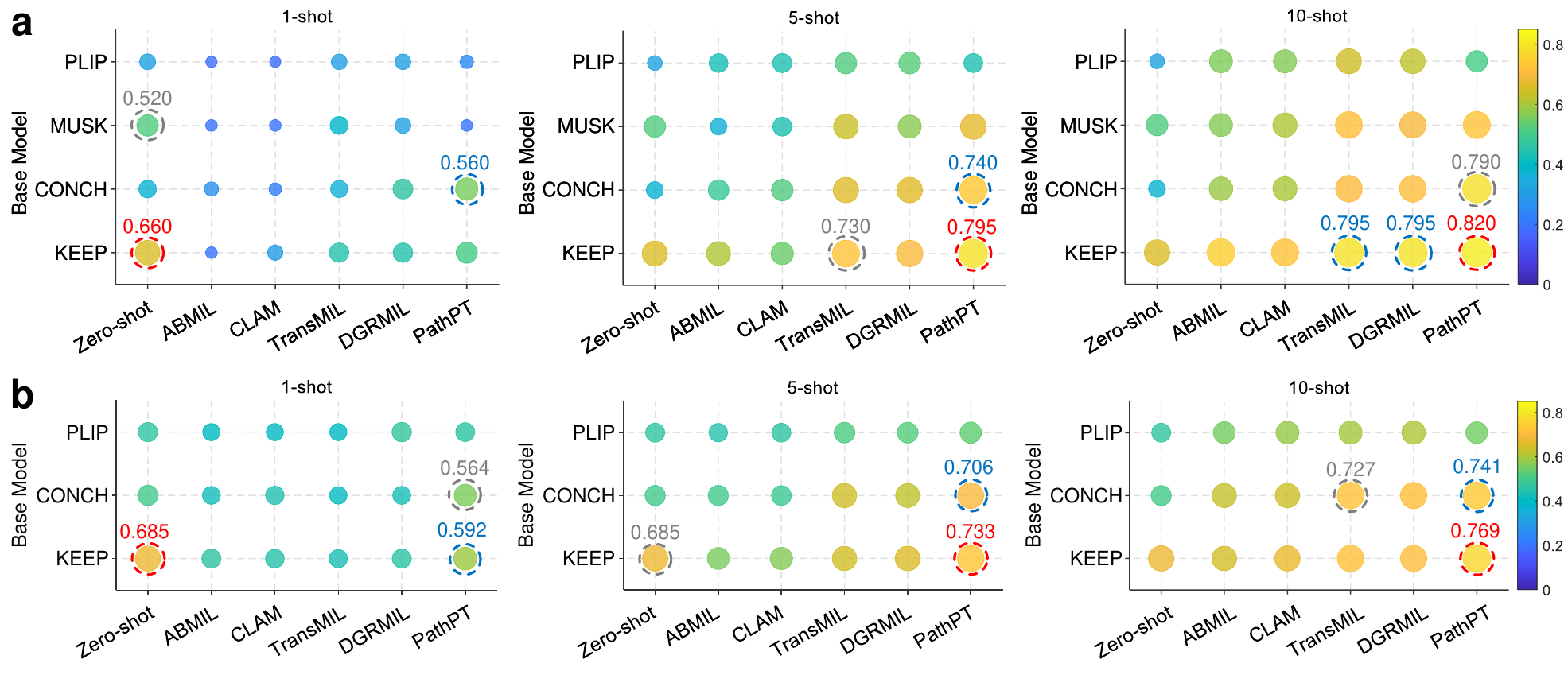}
  \caption{\textbf{Common cancer subtyping results}. 
    \textbf{a}. The performance comparison on UBC-OCEAN (kidney tumors with 5 subtypes) between PathPT and MIL baselines, including ABMIL, CLAM, TransMIL, and DGRMIL, using different base models (PLIP, MUSK, CONCH, and KEEP), in 1-shot, 5-shot, and 10-shot settings. Red and blue circles highlight the best, second-best, and third-best performance, respectively, across all base models and few-shot learning methods.
    \textbf{b}. Average performance on common cancer benchmark datasets from TCGA, including TCGA-BRAIN (brain tumors with 3 subtypes), TCGA-BRCA (breast cancer with 2 subtypes). MUSK is excluded because it was pretrained on TCGA. Red, blue, and gray circles highlight the best, second-best, and third-best performance, respectively, across all base models and few-shot learning methods.}
  \label{fig:result_common_sub}
  \vspace{-0.5cm}
\end{figure}

\subsection{Extending to Common Cancer Subtyping}

We further evaluated PathPT on common cancer subtyping using three public datasets: UBC-OCEAN~\cite{farahani2022ubc,asadi2024ubc2} (ovarian cancer), TCGA-BRCA (breast carcinoma), and TCGA-BRAIN (brain tumors).  
UBC-OCEAN includes \textbf{five} subtypes—ovarian clear cell carcinoma (35 WSIs), ovarian endometrioid carcinoma (35), high-grade serous ovarian carcinoma (35), low-grade serous ovarian carcinoma (35), and ovarian mucinous carcinoma (35).  
TCGA-BRCA comprises \textbf{two} subtypes—invasive ductal carcinoma (75) and invasive lobular carcinoma (75).  
TCGA-BRAIN encompasses \textbf{three} subtypes—glioblastoma (75), astrocytoma (75), and oligodendroglioma (73). Detialed information on these datasets can be found in Supplementary Table~\ref{tab:data_common}.

Overall performance comparisons are summarized in Figure~\ref{fig:result_common_sub}a (UBC-OCEAN) and Figure~\ref{fig:result_common_sub}b (TCGA), with detailed results in Supplementary Figure~\ref{fig:result_common_supp} and Supplementary Table~\ref{tab:ubc}-\ref{tab:brain}.
Rankings were consistent across datasets: KEEP provided the strongest backbone (zero-shot balanced accuracy of 0.660 on UBC and 0.685 on TCGA), followed by CONCH, MUSK, and PLIP. Among few-shot methods, PathPT paired with strong backbones achieved the highest performance, reaching 0.820 on UBC and 0.769 on TCGA in the 10-shot setting. DGRMIL and TransMIL became competitive at higher shots, whereas ABMIL was consistently weaker. The cancerous grounding performance and visualization of exemplary WSIs, shown in Supplementary Figure~\ref{fig:common_dice} and Supplementary Figure~\ref{fig:app_visual_sub_com1}-\ref{fig:app_visual_sub_com2}, confirms that PathPT significant improves tumor region grounding ability of vision-language foundation models.

\subsection{PathPT Enhances Cancerous Region Segmentation}

We next assessed PathPT on cancerous region segmentation using CAMELYON16~\cite{bejnordi2017camelyon16}, PANDA~\cite{bulten2022panda}, and AGGC22~\cite{huo2024aggc22}. For each benchmark, 15 WSIs with expert-provided masks were randomly sampled for training, with the remainder allocated for testing. Few-shot experiments were conducted with 1, 5, or 10 WSIs, each repeated 10 times for robustness. Unlike classification tasks, segmentation training used tile-level labels derived directly from the ground-truth masks. 

\begin{figure}[!t]
  \centering
  \includegraphics[width=\textwidth]{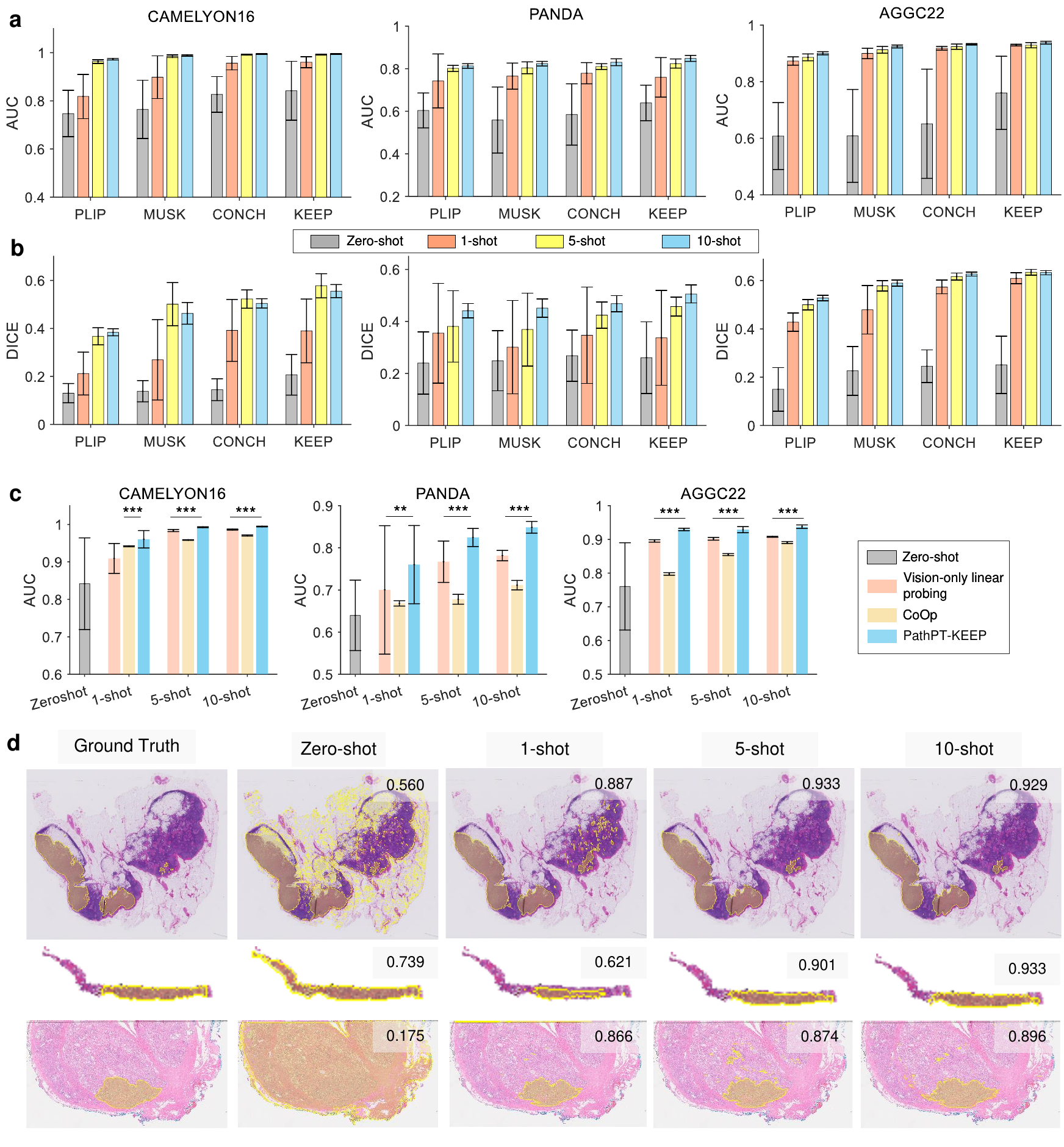}
  \caption{\textbf{Cancer region segmentation results}. 
    The AUC (\textbf{a}) and DICE (\textbf{b}) performance of PathPT on three cancerous region segmentation benchmarks using different base models.
  \textbf{c}. The performance comparison between PathPT-KEEP and baseline models, including vision linear probing and CoOp, respectively. A paired t-test assesses the statistical significance between the performance distributions. ** denotes P < 0.05, and *** denotes P < 0.005. 
  \textbf{d}. Representative whole slide images (WSIs) from three datasets are presented: CAMELYON16 (top row), PANDA (middle row), and AGGC22 (bottom row). Each row displays both ground truth segmentation masks and corresponding model predictions. The DICE score for each prediction is quantitatively indicated in the top-right corner of the respective image. The yellow masks denote tumor regions, either in ground truth or predictions. More segmentation visualizations can be found in Supplementary Figure~\ref{fig:app_visual_seg}.}
  \label{fig:result_seg}
\end{figure}

Results are presented in Figure~\ref{fig:result_seg}. Across all benchmarks, PathPT consistently improved segmentation performance as the number of training samples increased, demonstrating that the framework generalizes effectively to pixel-level tasks and benefits from the inductive bias of vision–language pretraining.

It can be observed that across all three datasets, few-shot PathPT based on different base models consistently achieves significant performance improvements over the zero-shot setting, indicating that fine-grained mask supervision can effectively enhance PathPT's tile-level classification capability. To further investigate the source of PathPT's effectiveness, we conducted two ablation studies on the KEEP base model: (1) vision-only linear probing, and (2) optimizing only the learnable prompt token (CoOp~\cite{zhou2022coop}). As shown in Figure~\ref{fig:result_seg}c, PathPT outperforms both linear probing and CoOp across all settings, demonstrating the advantage of combining learnable prompts with spatial visual context. Representative segmentation results on WSI instances are visualized in Figure~\ref{fig:result_seg}d, with more visualizations in Supplementary Figure~\ref{fig:app_visual_seg}.

\section{Discussion}

PathPT represents a shift in computational pathology by reframing how vision–language foundation models can be harnessed for rare cancer diagnosis. 
Our comprehensive benchmarking across eight rare cancer datasets (spanning 56 subtypes and 2,920 WSIs) and three common cancer datasets demonstrates PathPT's superior performance over traditional MIL frameworks. Notably, PathPT–KEEP achieved 0.679 balanced accuracy on EBRAINS under 10-shot settings, while delivering gains of more than 20 percentage points for pediatric cancers and showing strong performance in cancerous region segmentation tasks. These results establish a new benchmark for few-shot learning in rare cancer subtyping.

The core innovation of PathPT lies in moving beyond the aggregation-centric paradigm that dominates existing MIL approaches. Conventional methods treat foundation models as static feature generators and focus on learning optimal feature aggregation. Our findings highlight that solving misalignment between pre-trained representations and domain-specific semantics can also bring large improvements for rare cancer diagnosis.
By introducing learnable prompts that operate as adaptive queries, PathPT dynamically extracts the most relevant cross-modal knowledge, uncovering semantic relationships that handcrafted prompts cannot capture.

This design has critical implications for interpretability. Traditional MIL pipelines often obscure the contribution of individual patches through complex attention-weighted aggregation, limiting transparency and complicating clinical validation. In contrast, PathPT maintains tile-level grounding via zero-shot labeling, preserving a direct spatial link between predictions and tissue regions. 
This capability enables pathologists to cross-check AI outputs against their expertise, a prerequisite for trust in applications where unfamiliar morphological patterns may arise.

PathPT also addresses a major barrier to clinical deployment: adaptability. Its parameter-efficient prompt mechanism supports rapid customization to new or emerging subtypes without requiring large-scale data or computational resources. This is particularly relevant as cancer classification is inherently open-set, continuously expanding with the discovery of novel molecular and histopathological subtypes. By allowing minimal yet targeted adaptation, PathPT ensures diagnostic systems can remain aligned with evolving clinical standards while remaining feasible for deployment in diverse healthcare settings.

Beyond pathology, these findings suggest a broader research trajectory: foundation model adaptation strategies that prioritize semantic alignment over architectural complexity. 
The demonstration that cross-modal reasoning can be unlocked through prompt optimization—rather than extensive fine-tuning—provides a template for developing resource-efficient, interpretable AI systems in other medical fields where expertise is scarce and diagnostic accuracy is critical.

\section{Methods}

\subsection{Overview on Vision-language Pathology Foundation Models}
Pathology vision-language foundation models are typically composed of a tile-level pathology vision encoder $\Phi_{v}$ that encodes each tile image, into a visual representation $\mathbf{v} = \Phi_v(x)\in\mathbb{R}^d$, and a paired text encoder $\Phi_{t}$ that encodes a group of manually defined prompts for $N$ classes, $\mathbf{c} = \{c_1, c_2, ..., c_N\}$, into text representations $\mathbf{t}_i = \Phi_{t}(c_i)\in\mathbb{R}^{N\times d}$. 
The tile's category can be readout by taking the highest cosine similarity between the visual representation and the group of text representations:
\begin{equation}
    \hat{y} = \argmax_{i} \left< \mathbf{v},\mathbf{t}_i \right>
\end{equation}
where $\left< \cdot \right>$ denotes the cosine similarity between tile image embeddings and prompt embeddings.
When applied to whole slide diagnosis, the WSI is firstly divided into small tiles $X = \{x_1, x_2, ..., x_M \}$, with each tile independently classified by the vision-language representation similarities. The final WSI-level diagnosis is then aggregated by integrating tile-level predictions, typically through majority voting~\cite{zhou2024keep} or Top-k pooling~\cite{lu2024visual}.
Although these models demonstrate substantial advances in common cancer subtyping, their zero-shot capabilities for subtyping rare cancers are still insufficient to meet clinical deployment standards.

\subsection{Few-shot Cancer Subtyping}

Few-shot learning aims to adapt models to new tasks using only a limited number of labeled training examples. This paradigm is particularly crucial for rare cancer subtyping, where acquiring sufficient annotated samples is often challenging due to the low prevalence of these malignancies.
In this setting, there is a limited training set $\mathcal{D}_{\text{train}} = \{(\mathbf{X}_i, y_i), i=1,2,..., N\}$, where $N$ is small, each whole slide~($\mathbf{X}_i$) contains $M_i$ patches, {\em i.e.}, $\mathbf{X}_i = \{\mathbf{x}_{ij} | j=1,2,...,M_i\}$, with slide-level subtype label $y_i \in \{1, 2, \ldots, C\}$ for $C$ cancer subtypes. 
Each patch $\mathbf{x}_{ij}$ is processed by a pre-trained tile-level foundation model to extract visual features $\mathbf{v}_{ij} \in \mathbb{R}^{d}$. Traditional MIL frameworks typically learn to aggregate tile-level visual features into slide-level predictions:
\begin{equation}
\hat{y}_i = \mathcal{F}_{\text{MIL}}(\{\mathbf{v}_{ij}\}_{j=1}^{M_i}; \boldsymbol{\theta}_v)
\end{equation}
where $\hat{y}_{i}$ suggests the prediction of the $i$-th WSI. $\boldsymbol{\theta}_v$ suggests the weights of MIL models. This framework rely solely on visual features of vision-language models and lacks spatial granularity from slide-level supervision, limiting its ability to capture fine-grained patterns for rare cancer subtyping.

In this paper, we propose \textbf{PathPT}, a novel framework for tile-level classification within WSIs using spatially-aware visual aggregation and task-adaptive prompt tuning, which can be formulated by:
\begin{equation}
    \hat{y}_{ij} = \mathcal{F}_{\text{PathPT}}(\mathbf{v}_{ij}; \boldsymbol{\theta}_v; \boldsymbol{\theta}_t)
\end{equation}
where $\hat{y}_{ij}$ suggests the prediction of the $j$-th tile in the $i$-th WSI. 
$\boldsymbol{\theta}_v$ denotes the weights of the vision branch for spatially-aware visual aggregation. $\boldsymbol{\theta}_t$ represents learnable text tokens for task-adaptive prompt tuning. The detailed architecture and methodology of PathPT are present below.

\subsection{Model Architecture of PathPT}
The overall architecture of PathPT is shown in Figure~\ref{fig:teaser}b, where both vision and text encoders are initialized from pathology foundation models and remain frozen during training. For the vision branch, after extracting the original tile-level features $\mathbf{v}_i = \Phi_{v}(x_i)$ for all tiles in a WSI, these features are spatially ordered according to their coordinates to form a sequence. This spatially arranged sequence is then processed through a spatially-aware module consisting of two components:
i) Local interaction block: A residual convolutional module with parallel kernels of sizes $3\times3$, $5\times5$ and $7\times7$, capturing local feature interactions between adjacent tiles within sliding windows across the WSI grid.
ii) Global interaction Layer: A transformer encoder layer employing self-attention mechanisms to model long-range dependencies and global contextual relationships among all tiles in the sequence:
\begin{equation}
    \bar{\mathbf{v}}_1, \bar{\mathbf{v}}_2,..., \bar{\mathbf{v}}_M = \Psi(\Phi_v(x_1), \Phi_v(x_2), ..., \Phi_v(x_M))
\end{equation}
where $\Psi(\cdot)$ suggest the spatial-aware module.

For the text branch, inspired by CoOp~\cite{zhou2022coop}, we propose to learn a context prompt for each class, which is given by: 
\begin{equation}
    \bar{\mathbf{c}}=[T]_1,[T]_2,...,[T]_K,[CLASS]
\end{equation}
where $[T]_i\in\mathbb{R}^{d\times 1}$ denotes a vector with the same dimension as the token embedding, $K$ is a hyperparameter specifying the number of learnable tokens. $[CLASS]$ suggests the token embeddings of different class names. As a result, the learnable context embeddings can be formulated by $\{\bar{\mathbf{c}}_1, \bar{\mathbf{c}}_2, ..., \bar{\mathbf{c}}_N\}$, where $N$ suggests the number of classes. Correspondingly, the prediction probability of each tile image can be calculated by:
\begin{equation}
    p\left( y=j|x_i \right) =\frac{\exp \left( \left< \Phi _t\left( \bar{\mathbf{c}}_j \right) ,\bar{\mathbf{v}}_i \right> /\tau \right)}{\sum_{j=1}^N{\exp \left( \left< \Phi _t\left( \bar{\mathbf{c}}_j \right) ,\bar{\mathbf{v}}_i \right> /\tau \right)}}
\end{equation}
where $\bar{\mathbf{v}}_i$ and $\bar{\mathbf{c}}_j$ suggest the visual representation of the $i$-th tile image and the learnable context of $j$-th class, respectively. $\left< \cdot \right>$ denotes the cosine similarity between tile image and context representations. $\tau$ is a temperature parameter. 

\subsection{Tile-level Supervision}

\textbf{Cancer Subtyping Tasks.}
Traditional multi-instance learning approaches leverage slide-level labels to perform WSI subtyping tasks, yet they suffer from coarse-grained supervision during training and limited interpretability during inference. In contrast, our PathPT framework exploits zero-shot tile-level predictions as fine-grained pseudo-labels to guide the prompt tuning process. To this end, we develop an effective approach that generates tile-level labels for slide images using only WSI-level labels.
\begin{itemize}
    \item \textbf{Manual Prompt Selection.} To generate high-quality tile-level labels, we first construct a diverse prompt pool by randomly combining multiple predefined prompt templates with different WSI category names, yielding 200 distinct prompt groups in total. Each prompt template follows standard natural language patterns, such as ``A histopathological image of \{category\}'' or ``An image showing \{category\}'', where \{category\} is substituted with specific tumor pathological subtype names. The complete set of templates is consistent with CONCH~\cite{lu2024visual}. Subsequently, we evaluate all 200 prompt groups by ranking their WSI classification performance on the training dataset and select the top-performing 100 groups for prompt embedding generation. Rather than utilizing individual prompts independently, we employ mean pooling to aggregate the embeddings from these 100 selected prompt groups into a unified representation.
    \item  \textbf{Zero-shot Tile Labeling.} After obtaining the text prompt embeddings, we compute the similarity between these embeddings and each tile to generate label predictions for individual tiles. Rather than directly using all tile-level predictions as pseudo-labels, we selectively retain only those predictions that are normal or consistent with the corresponding WSI-level label. Specifically, for a WSI labeled as subtype A, we only consider tiles predicted as either normal tissue or subtype A as valid pseudo-labels, while discarding predictions that conflict with the WSI-level ground truth (e.g., tiles predicted as other subtypes). This selective pseudo-labeling strategy ensures that the tile-level supervision remains coherent with the slide-level annotation.
\end{itemize}

\textbf{Cancer Region Segmentation Tasks.} For WSI-level cancer region segmentation tasks. The tile-level labels are generated from the ground truth cancerous masks.

\subsection{Training Loss Function}
\textbf{Labeled Tile Loss.} A balanced cross-entropy loss $\mathcal{L}_{labeled}$ is applied to tiles with known labels, encompassing ground truth labels for cancer region segmentation tasks, as well as predicted labels for cancer subtyping tasks.

\textbf{Unlabeled Tile Loss.} For cancer subtyping tasks, in addition to the labeled tile loss, we introduce a candidate loss for tiles that are not assigned pseudo-labels. We assume that these unlabeled tiles belong to either normal tissue or the current WSI's subtype category. Specifically, for an unlabeled tile from a WSI of subtype $i$, we constrain the sum of its logits corresponding to the normal class (noted by $0$) and subtype $i$ to equal $1$. This candidate loss can be formulated as:
\begin{equation}\label{eq5}
    \mathcal{L}_{unlabeled}= -\text{log}(p(y=0|x) + p(y=i|x))
\end{equation}
\textbf{Pseudo-label Loss.} Moreover, we incorporate pseudo-label loss starting from the $10$-th epoch. At this stage, we employ the current model to generate labels for unlabeled tiles and subsequently construct a balanced cross-entropy loss using these pseudo-labels. We enable this loss only for CONCH and KEEP backbones; for MUSK and PLIP, we disable it due to unstable training dynamics and inconsistent performance.

\subsection{Difference between MIL and PathPT}
MIL frameworks aggregate tile-level features into a slide-level representation via attention and perform WSI classification.
However, MIL approaches typically require large numbers of training samples to learn effective attention mechanisms and feature aggregations, making them less suitable for rare cancer scenarios where data availability is inherently limited. Moreover, MIL lacks explicit tile-level supervision, limiting its ability to localize or interpret tumor regions.
In contrast, PathPT enables effective learning with minimal training samples by leveraging the pre-trained knowledge of VL models. PathPT combines WSI labels with VL model tile-level predictions to generate pseudo-labels for each tile, enabling fine-grained, tile-level training. As a result, PathPT provides detailed tile-level predictions at inference, supporting both subtype identification and tumor localization within WSIs for enhanced interpretability.

\subsection{Model Training Details} 
We trained the model for a total of 20 epochs with a learning rate of 1e-4 and a 2-epoch warm-up period. Each experiment was repeated 10 times with different random selections of training samples to ensure robustness. The number of learnable tokens was set to 32, and their embeddings were initialized using the token embeddings from manual prompts. For all subtyping tasks, the loss weights for labeled tile loss, unlabeled tile loss, and pseudo-label loss were set to 1:0.5:0.1, respectively. All experiments were conducted on a single NVIDIA GeForce RTX 4090 GPU.

\subsection{Evaluation Metrics}

For cancer subtyping tasks, we follow the same evaluation protocol as KEEP, employing the tumor-ratio approach to generate WSI-level subtype prediction and using balanced accuracy (BACC) to assess the performance:
\begin{equation}
    \text{BACC} = \frac{1}{C} \sum_{i=1}^{C} \frac{\text{TP}_i}{\text{TP}_i + \text{FN}_i}
\end{equation}
where $C$ is the number of classes, $\text{TP}_i$ and $\text{FN}_i$ are the number of true positives and false negatives for the $i$-th subtype, respectively.

For WSI-level cancer region segmentation tasks, we utilize the area under the curve (AUC) and DICE score to evaluate different models:
\begin{equation}
        \text{AUC} = \int_{0}^{1} \text{TPR}(f) \cdot \text{FPR}'(f) \, df   
\end{equation}
\begin{equation}
    \text{DICE} = \frac{2 \cdot |X \cap Y|}{|X| + |Y|}
\end{equation}
where $\text{TPR}(f)$ is the true positive rate and $\text{FPR}(f)$ is the false positive rate at threshold $f$. $X$ and $Y$ are the predicted and ground truth segmentation masks, respectively.

\section{Data Availability}
For cancer subtyping, test datasets of TCGA-BRCA, TCGA-BRAIN (including TCGA-GBM, TCGA-LGG), TCGA-SARC, TCGA-UCS, TCGA-THYM are available in TCGA~(\href{https://portal.gdc.cancer.gov/}{https://portal.gdc.cancer.gov/}). Other datasets for cancer subtyping are available in EBRAINS~(\href{https://data-proxy.ebrains.eu/datasets/}{https://data-proxy.ebrains.eu/datasets/}), and UBC-OCEAN ~(\href{https://www.kaggle.com/competitions/UBC-OCEAN/}{https://www.kaggle.com/competitions/UBC-OCEAN/}). Test datasets for cancer region segmentation are available in CAMELYON16~(\href{https://camelyon16.grand-challenge.org/}{https://camelyon16.grand-challenge.org/}), PANDA~(\href{https://panda.grand-challenge.org/data/}{https://panda.grand-challenge.org/data/}), and AGGC22~(\href{https://aggc22.grand-challenge.org/}{https://aggc22.grand-challenge.org/}).
For rare pediatric cancers, all WSIs and clinical information from Xinhua Hospital represent confidential patient data that cannot be publicly released.

\section{Code Availablity}
The source codes for PathPT are available at~\href{https://github.com/MAGIC-AI4Med/PathPT}{https://github.com/MAGIC-AI4Med/PathPT}.


\clearpage

\bibliographystyle{sn-mathphys} 
\bibliography{sn-bibliography} 

\clearpage

\appendix

\renewcommand{\thefigure}{S\arabic{figure}}
\renewcommand{\figurename}{Supplementary Figure}

\renewcommand{\thetable}{S\arabic{table}}
\renewcommand{\tablename}{Supplementary Table}
\setcounter{figure}{0}
\setcounter{table}{0}
 
\newpage

\begin{figure}[!t]
  \centering
  \includegraphics[width=\textwidth]{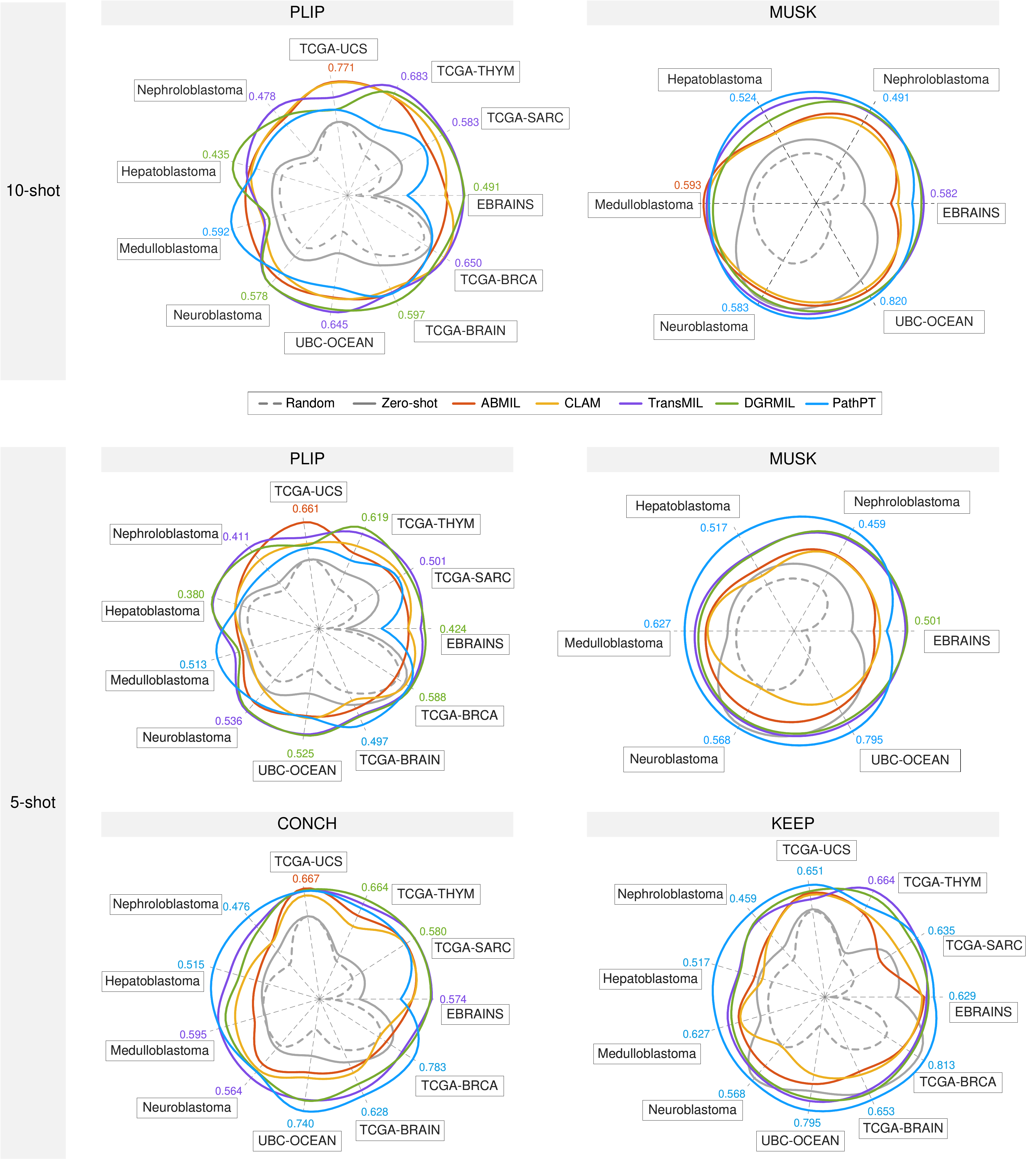}
  \caption{\textbf{Overall results}. 
  The average performance of different approaches across all subtyping tasks using diverse base models. The values denote the highest performance, and the colors indicate the corresponding best-performing methods. PathPT-MUSK and PathPT-KEEP achieve the best 5-shot performance on 5/6 and 10/11 cancer subtyping benchmarks, respectively.}
  \label{fig:teaser_supp}
\end{figure}

\begin{figure}[!t]
  \centering
  \includegraphics[width=\textwidth]{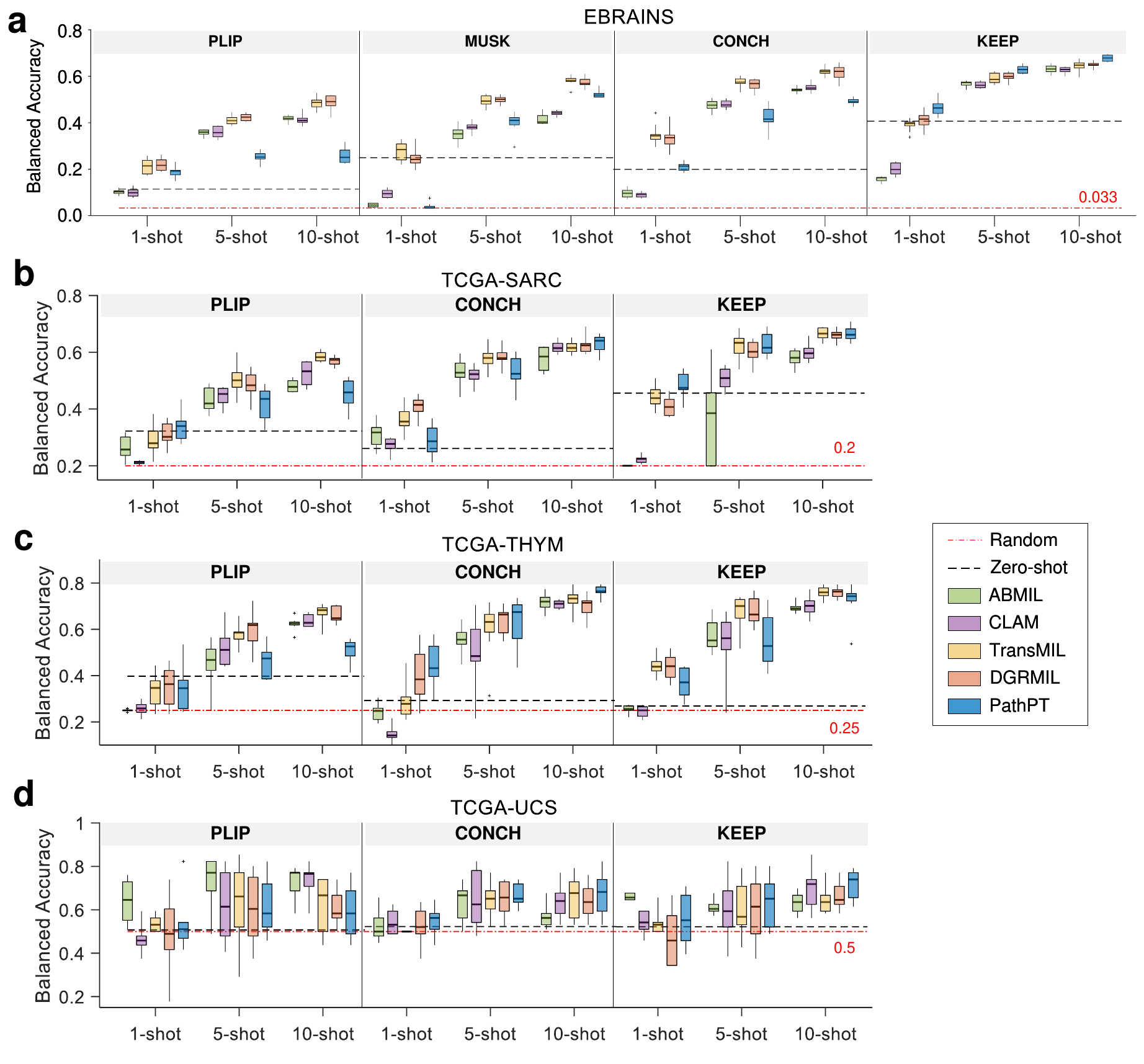}
  \caption{\textbf{Rare adult cancer subtyping results}. 
  The performance comparison on rare adult cancer benchmarks, including EBRAINS (brain tumors, 30 subtypes), TCGA-SARC (sarcoma,5 subtypes), TCGA-THYM (thymoma, 4 subtypes), and TCGA-UCS (Uterine carcinosarcoma, 2 subtypes), between PathPT and MIL baselines, including ABMIL, CLAM, TransMIL, and DGRMIL, using different base models (PLIP, MUSK, CONCH, and KEEP). Note that MUSK is excluded from the experiments on TCGA-SARC, TCGA-THYM, and TCGA-UCS, as it is pretrained on TCGA data. The box plots present the median, first, and third quartiles of 10-repeat experimental results.}
  \label{fig:result_rare_adult_supp}
\end{figure}

\begin{figure}[!t]
  \centering
  \includegraphics[width=\textwidth]{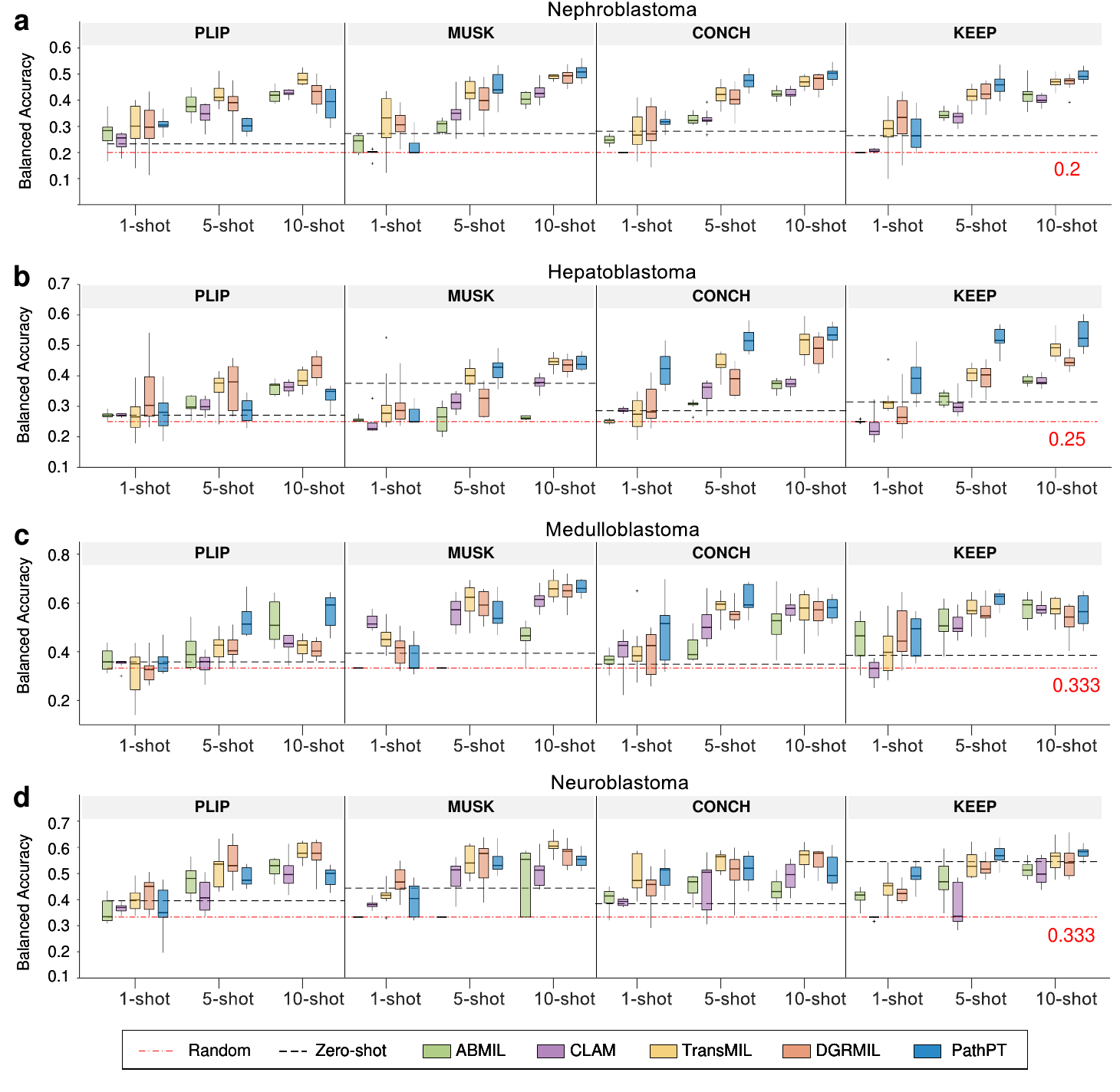}
  \caption{\textbf{Rare adult cancer subtyping results}. 
  The performance comparison on rare pediatric cancer subtyping tasks, including nephroblastoma (5 subtypes), hepatoblastoma(4 subtypes), medulloblastoma (3 subtypes), and neuroblastoma (3 subtypes). The box plots present the median, first, and third quartiles of 10-repeat experimental results.}
  \label{fig:result_rare_children_supp}
\end{figure}

\begin{figure}[!t]
  \centering
  \includegraphics[width=\textwidth]{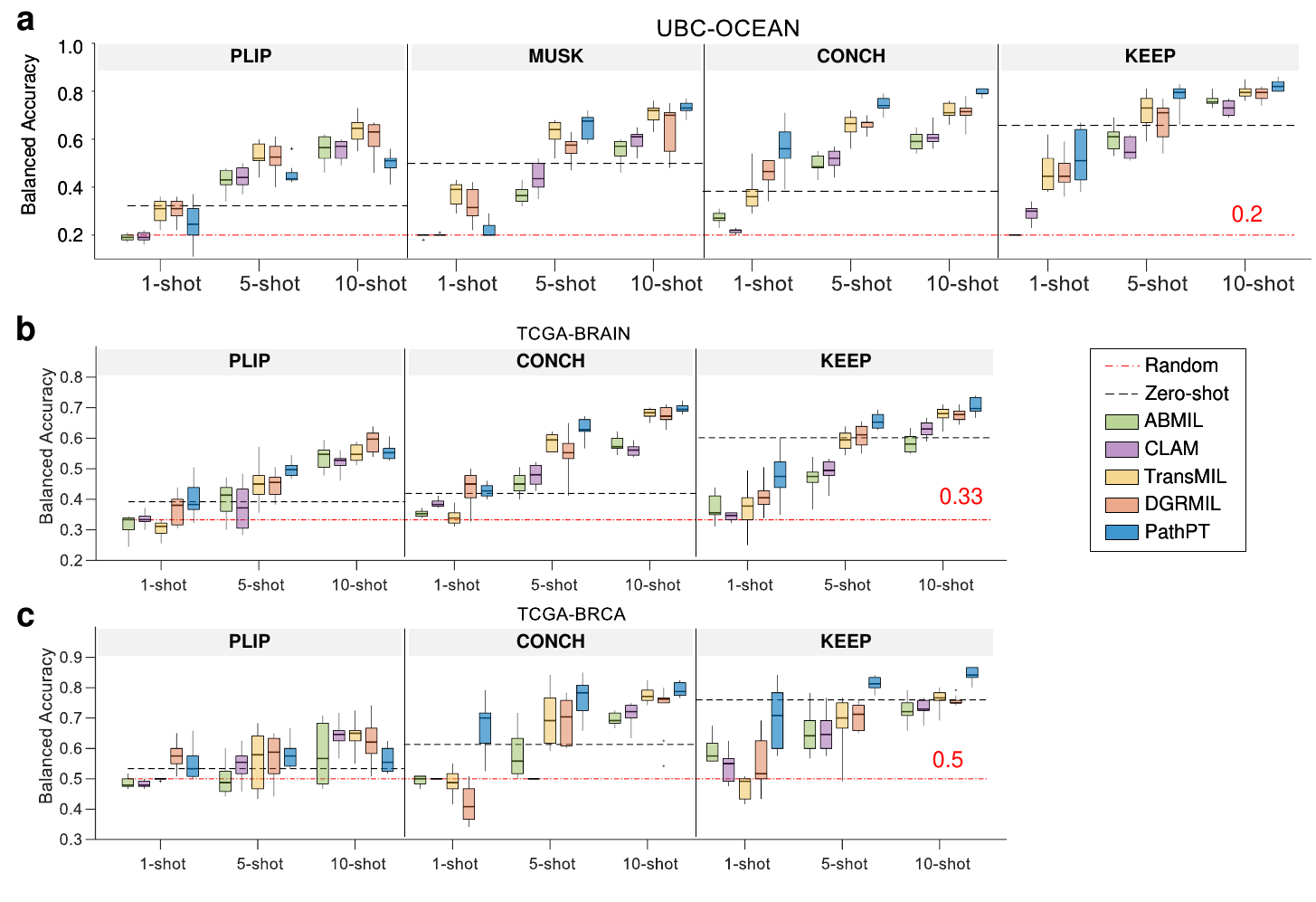}
  \caption{\textbf{Common cancer subtyping results}. 
  The performance comparison on common cancer subtyping tasks, including UBC-OCEAN (5 subtypes), TCGA-BRAIN (3 subtypes), and TCGA-BRCA (2 subtypes). Note that MUSK is excluded from the experiments on TCGA-BRAIN AND TCGA-BRCA, as it is pretrained on TCGA data. The box plots present the median, first, and third quartiles of 10-repeat experimental results.}
  \label{fig:result_common_supp}
\end{figure}

\begin{figure}[!t]
  \centering
  \includegraphics[width=\textwidth]{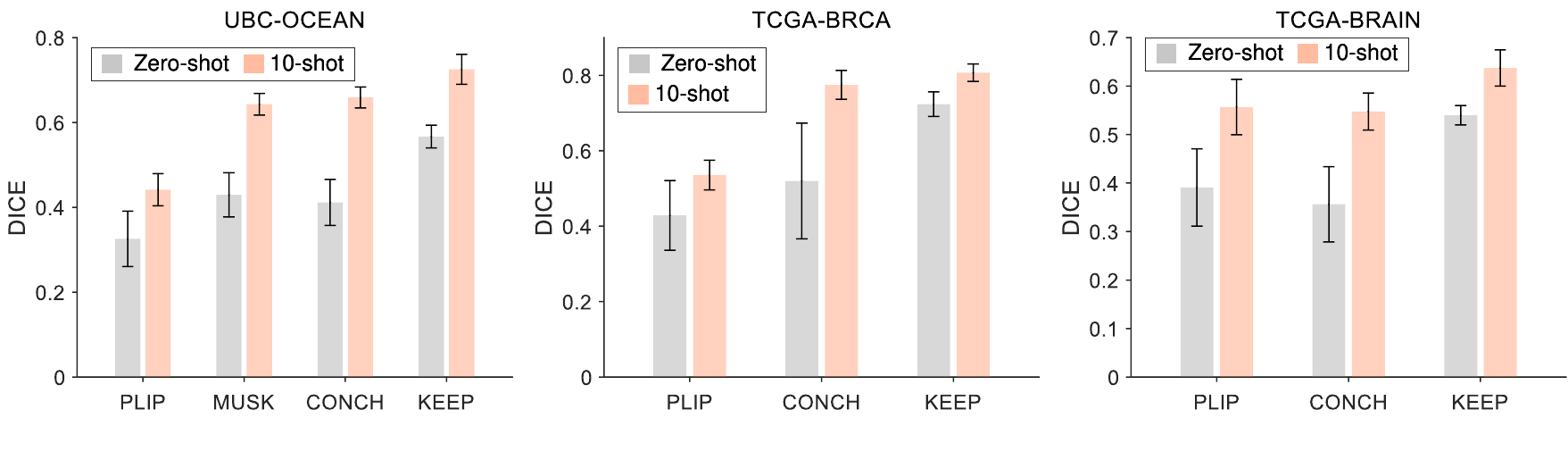}
  \caption{\textbf{Common cancer grounding performance}. 
  The Dice performance comparison on common cancer subtyping tasks, including UBC-OCEAN (5 subtypes), TCGA-BRAIN (3 subtypes), and TCGA-BRCA (2 subtypes). Note that MUSK is excluded from the experiments on TCGA-BRAIN AND TCGA-BRCA, as it is pretrained on TCGA data.}
  \label{fig:common_dice}
\end{figure}

\begin{figure}[!b]
    \centering
    \includegraphics[width=.8\textwidth]{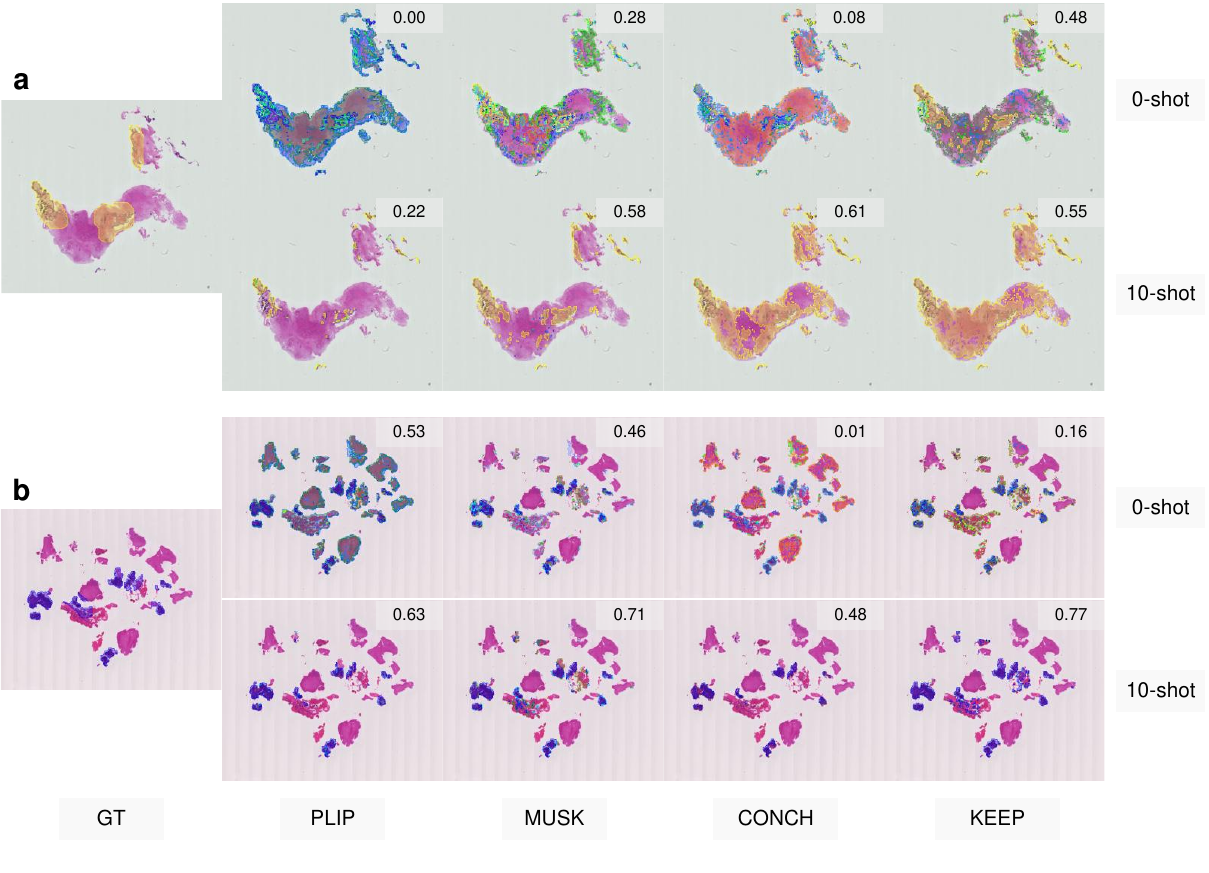}
    \caption{\textbf{Rare brain cancer grounding (EBRAINS).} 
        Representative WSIs from EBRAINS showing brain cancer subtypes: adamantinomatous craniopharyngioma (\textbf{a}) and anaplastic ependymoma (\textbf{b}). For each sub-figure, the two rows show 0-shot and 10-shot predictions respectively. Each row displays ground truth (\textbf{GT}) and corresponding model predictions from PLIP, MUSK, CONCH, and KEEP. Different tumor subtypes are marked with different colored masks. The DICE score is quantitatively indicated in the top-right corner of each prediction and is calculated only for the specific subtype corresponding to the ground truth mask. The colored masks denote tumor regions in both ground truth and predictions.
    }
    \label{fig:app_visual_sub_adult3}
\end{figure}

\begin{figure}[!t]
    \centering
    \includegraphics[width=.8\textwidth]{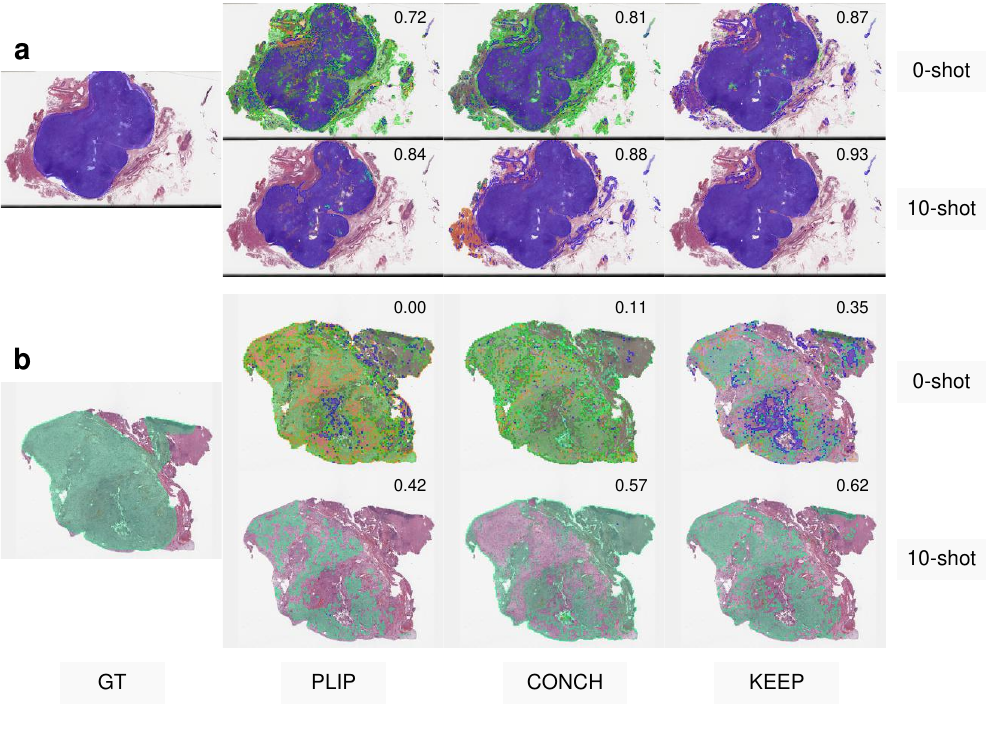}
    \caption{\textbf{Rare adult tumor grounding (TCGA-SARC, sarcoma).} 
        Representative WSIs from TCGA showing sarcoma subtypes: leiomyosarcoma (\textbf{a}) and malignant peripheral nerve sheath tumors (\textbf{b}). For each sub-figure, the two rows show 0-shot and 10-shot predictions respectively. Each row displays ground truth (\textbf{GT}) and corresponding model predictions from PLIP, CONCH, and KEEP. Different tumor subtypes are marked with different colored masks. The DICE score is quantitatively indicated in the top-right corner of each prediction and is calculated only for the specific subtype corresponding to the ground truth mask. The colored masks denote tumor regions in both ground truth and predictions. Due to the TCGA-specific pre-training of MUSK, we excluded results involving MUSK. 
    }
    \label{fig:app_visual_sub_adult2}
\end{figure}

\begin{figure}[!t]
    \centering
    \includegraphics[width=.8\textwidth]{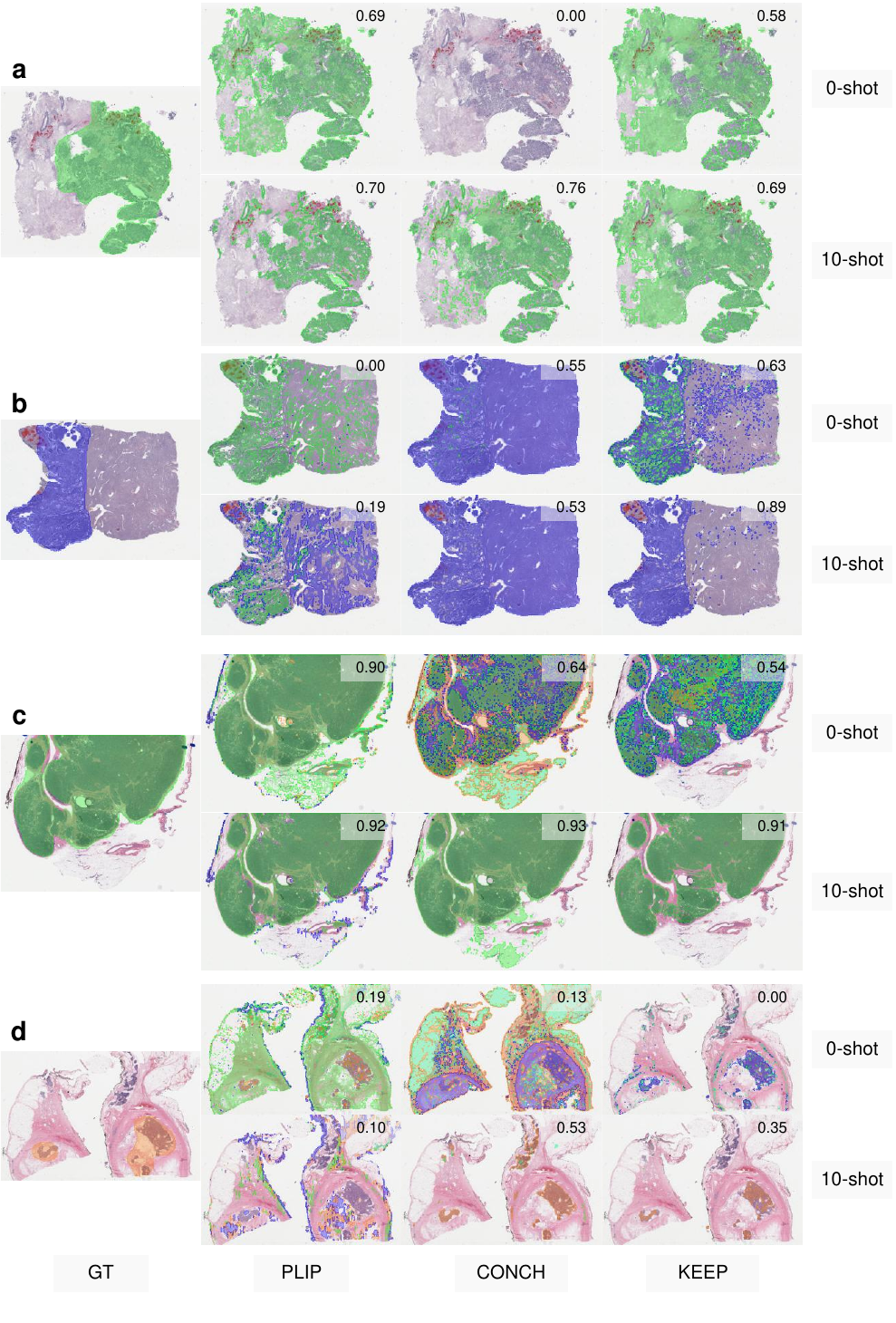}
    \caption{\textbf{Rare adult tumor grounding (TCGA-THYM, thymoma; TCGA-UCS, uterine carcinosarcoma).} 
        Representative WSIs from TCGA showing uterine carcinosarcoma subtypes: heterologous type (\textbf{a}) and homologous type (\textbf{b}), and thymoma subtypes: Type A (\textbf{c}) and Type B2 (\textbf{d}). For each sub-figure, the two rows show 0-shot and 10-shot predictions respectively. Each row displays ground truth (\textbf{GT}) and corresponding model predictions from PLIP, CONCH, and KEEP. Different tumor subtypes are marked with different colored masks. The DICE score is quantitatively indicated in the top-right corner of each prediction and is calculated only for the specific subtype corresponding to the ground truth mask. The colored masks denote tumor regions in both ground truth and predictions. Due to the TCGA-specific pre-training of MUSK, we excluded results involving MUSK. 
    }
    \label{fig:app_visual_sub_adult1}
\end{figure}

\begin{figure}[!t]
    \centering
    \includegraphics[width=\textwidth]{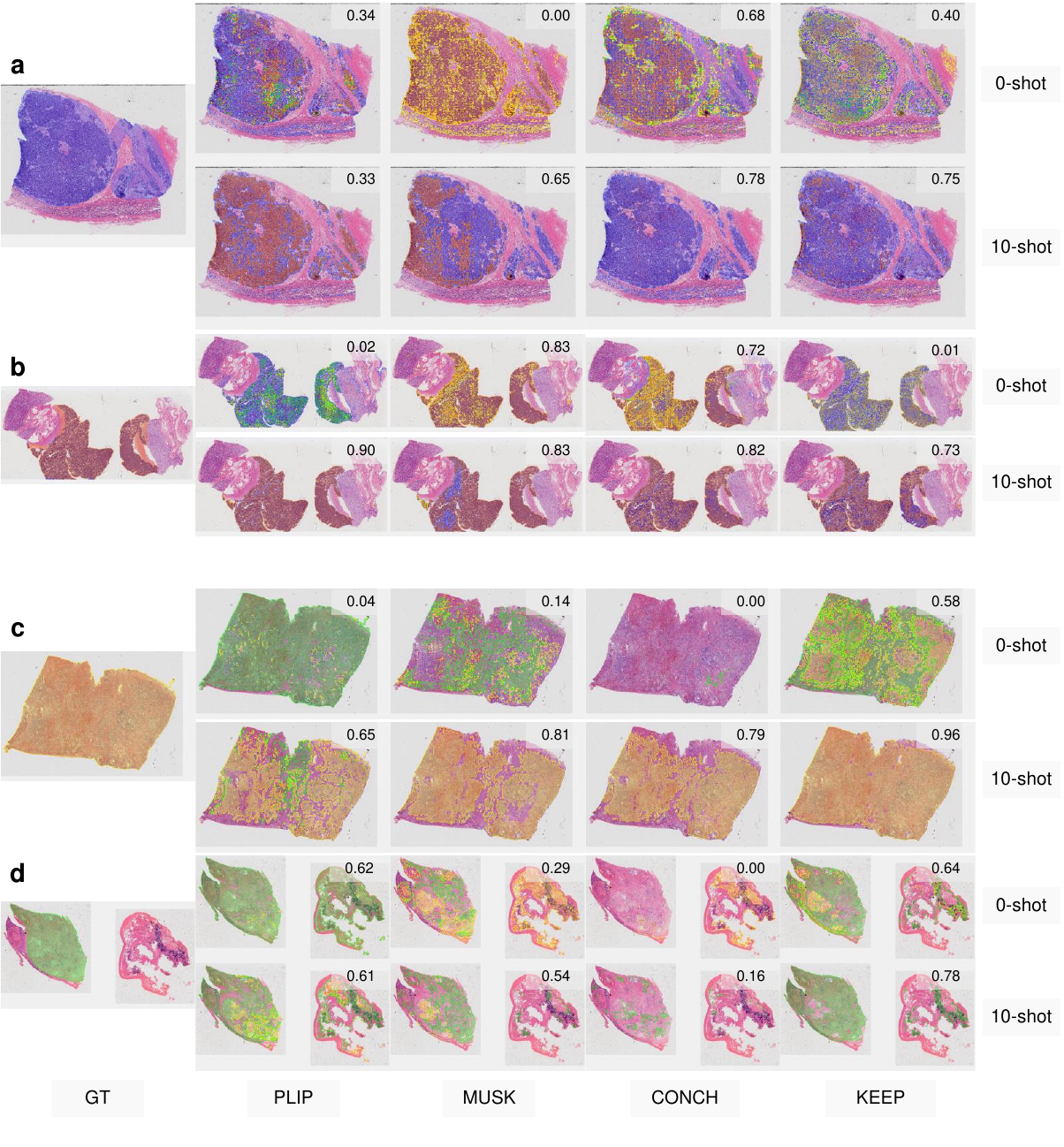}
    \caption{\textbf{Pediatric tumor grounding (Nephroblastoma \& Neuroblastoma).} 
        Representative WSIs from pediatric tumor datasets showing Nephroblastoma subtypes: Mixed blastemal, epithelial and stromal nephroblastoma (\textbf{a}) and Mixed blastemal and epithelial nephroblastoma (\textbf{b}), and Neuroblastoma subtypes: Differentiating neuroblastoma (\textbf{c}) and Ganglioneuroblastoma, intermixed (\textbf{d}). For each sub-figure, the two rows show 0-shot and 10-shot predictions respectively. Each row displays ground truth (\textbf{GT}) and corresponding model predictions from PLIP, MUSK, CONCH, and KEEP. Different tumor subtypes are marked with different colored masks. The DICE score is quantitatively indicated in the top-right corner of each prediction and is calculated only for the specific subtype corresponding to the ground truth mask. The colored masks denote tumor regions in both ground truth and predictions.
    }
    \label{fig:app_visual_sub_child1}
\end{figure}

\begin{figure}[!t]
    \centering
    \includegraphics[width=.9\textwidth]{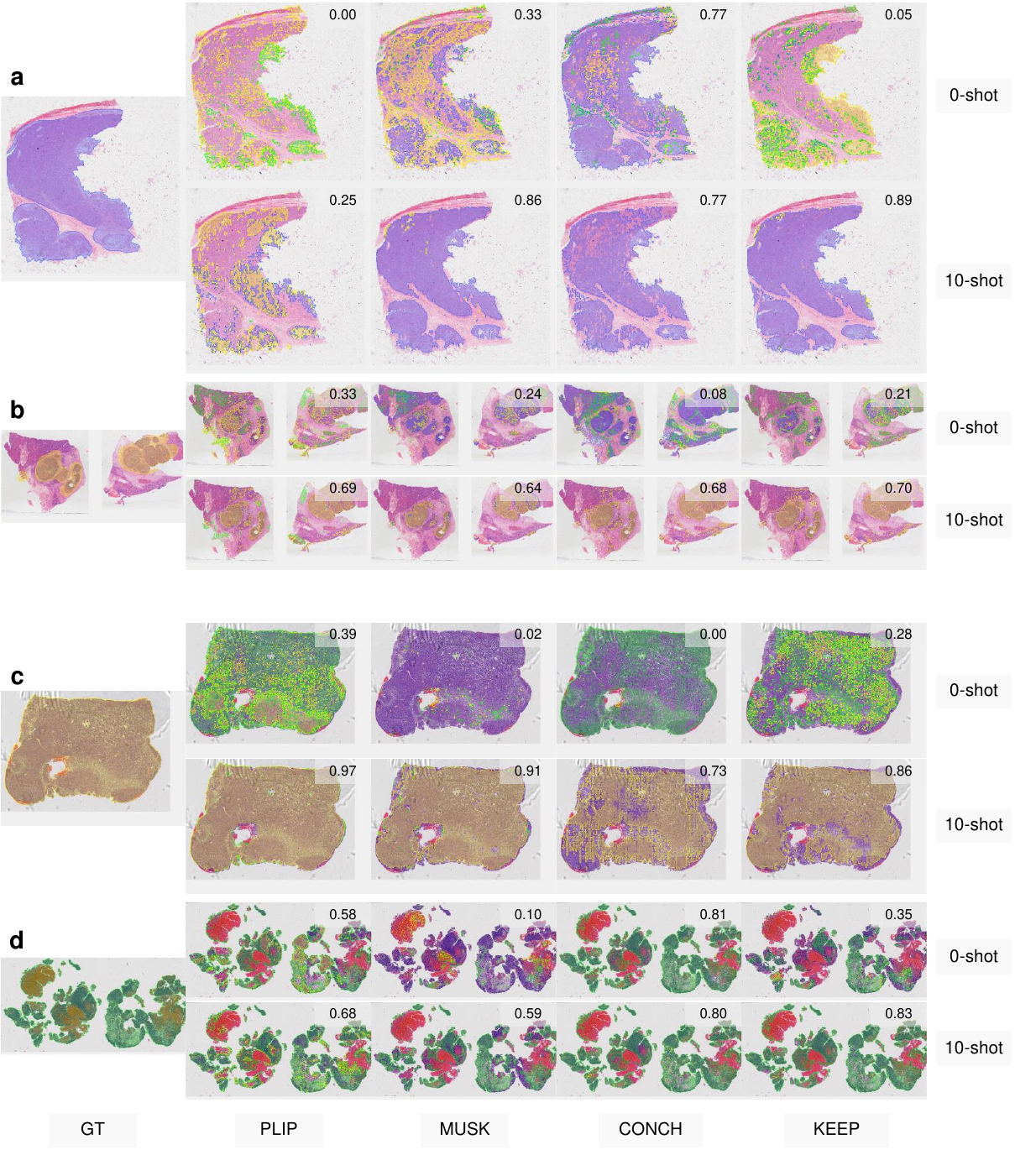}
    \caption{\textbf{Pediatric tumor grounding (Hepatoblastoma \& Medulloblastoma).} 
        Representative WSIs from pediatric tumor datasets showing Hepatoblastoma subtypes: Epithelial macrotrabecular pattern of hepatoblastoma (\textbf{a}) and Epithelial mixed fetal and embryonal hepatoblastoma (\textbf{b}), and Medulloblastoma subtypes: Desmoplastic nodular medulloblastoma (\textbf{c}) and Classic medulloblastoma (\textbf{d}). For each sub-figure, the two rows show 0-shot and 10-shot predictions respectively. Each row displays ground truth (\textbf{GT}) and corresponding model predictions from PLIP, MUSK, CONCH, and KEEP. Different tumor subtypes are marked with different colored masks. The DICE score is quantitatively indicated in the top-right corner of each prediction and is calculated only for the specific subtype corresponding to the ground truth mask. The colored masks denote tumor regions in both ground truth and predictions.
    }
    \label{fig:app_visual_sub_child2}
\end{figure}

\begin{figure}[!t]
    \centering
    \includegraphics[width=.75\textwidth]{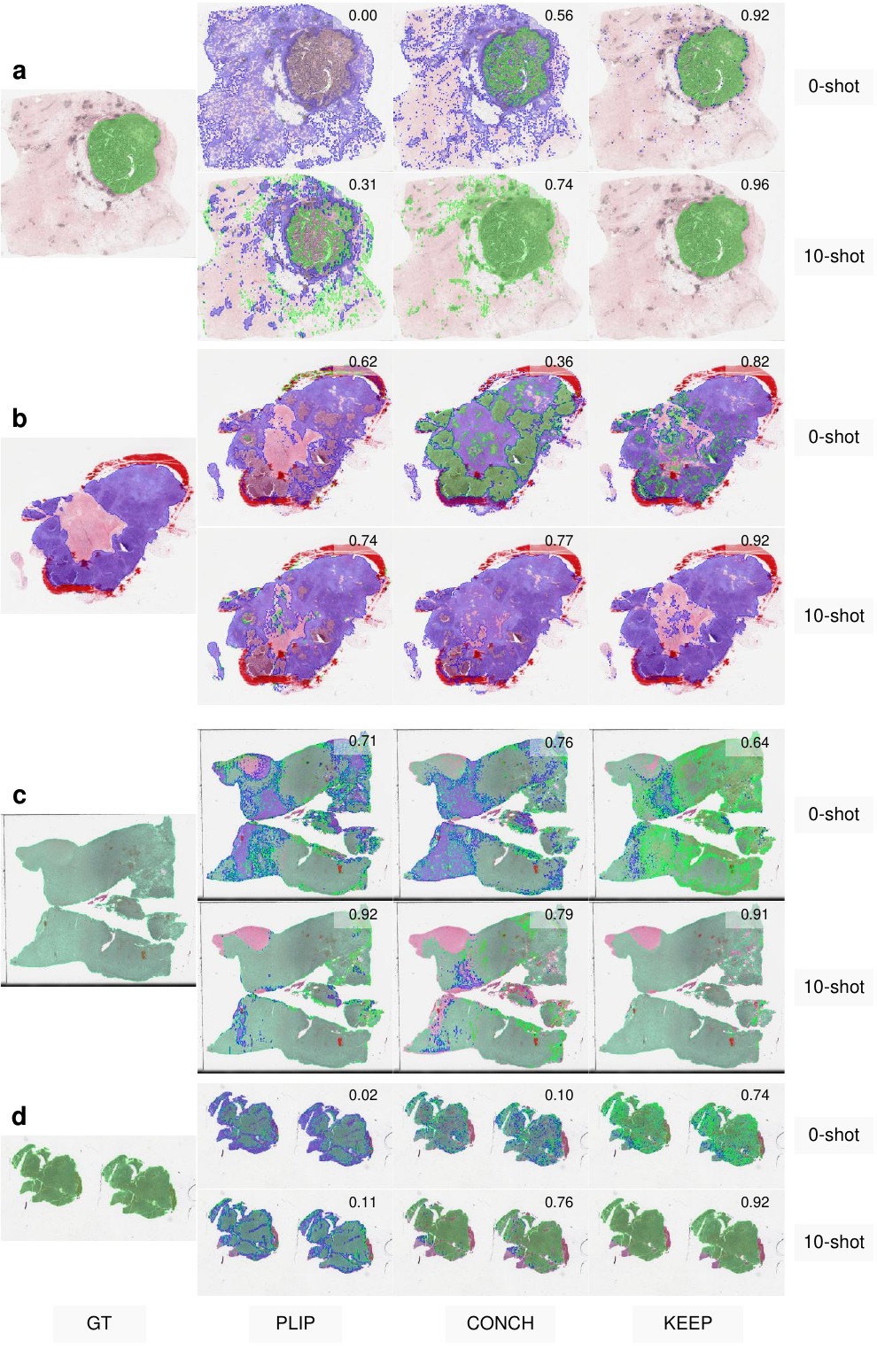}
    \caption{\textbf{Common cancer grounding (TCGA-BRCA, TCGA-BRAIN).} 
        Representative WSIs from TCGA showing invasive breast carcinom subtypes: invasive ductal carcinoma (\textbf{a}) and invasive lobular carcinoma (\textbf{b}), and brain tumor subtypes: oligodendroglioma (\textbf{c}) and glioblastoma (\textbf{d}). For each sub-figure, the two rows show 0-shot and 10-shot predictions respectively. Each row displays ground truth (\textbf{GT}) and corresponding model predictions from PLIP, CONCH, and KEEP. Different tumor subtypes are marked with different colored masks. The DICE score is quantitatively indicated in the top-right corner of each prediction and is calculated only for the specific subtype corresponding to the ground truth mask. The colored masks denote tumor regions in both ground truth and predictions. Due to the TCGA-specific pre-training of MUSK, we excluded results involving MUSK.
        }
    \label{fig:app_visual_sub_com1}
\end{figure}

\begin{figure}[!t]
    \centering
    \includegraphics[width=\textwidth]{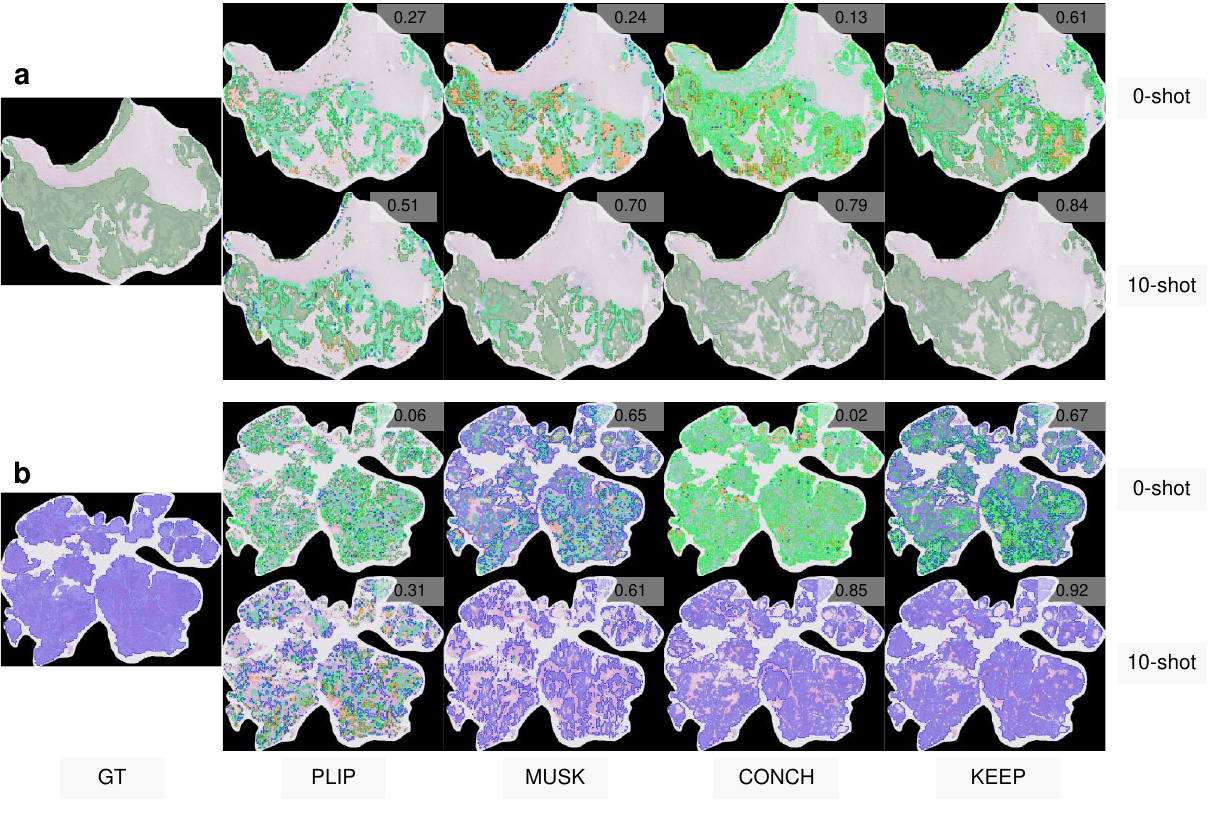}
    \caption{\textbf{Common cancer grounding (UBC-OCEAN).} 
        Representative WSIs from UBC-OCEAN showing ovarian cancer subtypes: ovarian mucinous carcinoma (\textbf{a}) and low-grade serous ovarian carcinoma (\textbf{b}). For each sub-figure, the two rows show 0-shot and 10-shot predictions respectively. Each row displays ground truth (\textbf{GT}) and corresponding model predictions from PLIP, MUSK, CONCH, and KEEP. Different tumor subtypes are marked with different colored masks. The DICE score is quantitatively indicated in the top-right corner of each prediction and is calculated only for the specific subtype corresponding to the ground truth mask. The colored masks denote tumor regions in both ground truth and predictions.
        }
    \label{fig:app_visual_sub_com2}
\end{figure}

\begin{figure}[!t]
    \centering
    \includegraphics[width=.9\textwidth]{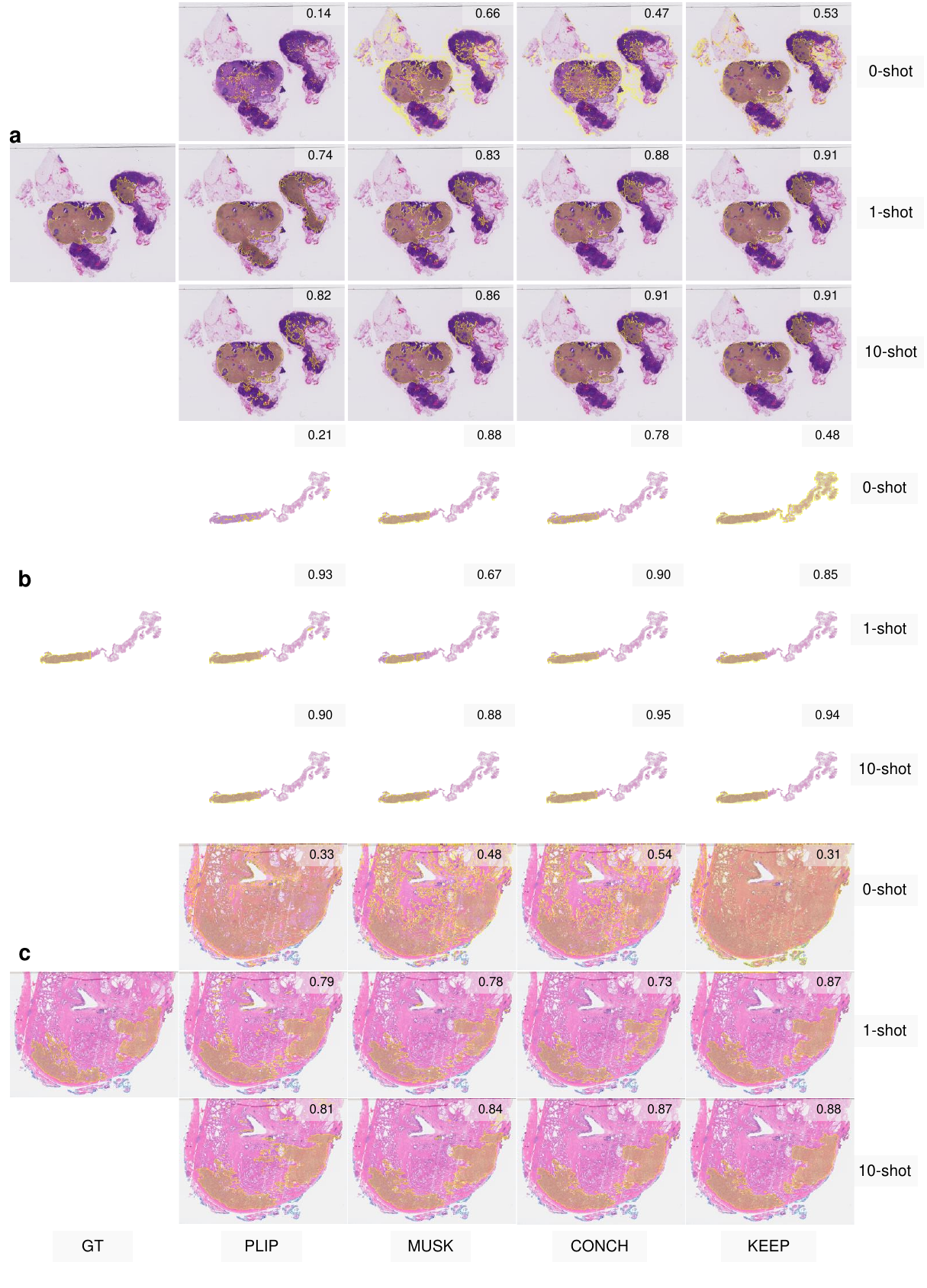}
    \caption{\textbf{Cancerous region segmentation.} 
        Representative WSIs from CAMELYON16 (\textbf{a}, top 3 rows), PANDA (\textbf{b}, middle 3 rows), and AGGC22 (\textbf{c}, bottom 3 rows). For each dataset, the three rows show zero-shot, 1-shot, and 10-shot predictions respectively. Each row displays both ground truth (\textbf{GT}) and corresponding model predictions, including PLIP, MUSK, CONCH, and KEEP. The DICE score is quantitatively indicated in the top-right corner of the respective image. The yellow masks denote tumor regions, either in ground truth or predictions.}
    \label{fig:app_visual_seg}
\end{figure}

\clearpage

\begin{table}[]
\small
\centering
\caption{\textbf{Experimental results on EBRAINS.} Performance comparison of PathPT and various MIL methods across different base models on the EBRAINS. Values represent the median of 10 runs, with Q1 and Q3 quartiles shown in parentheses. Bold indicates the best performance and \textcolor{blue}{blue} indicates the second-best performance.}
\scalebox{0.75}{
\setlength{\tabcolsep}{{5pt}}
\begin{tabular}{ccccccc}
\hline
                                                                                      &         & ABMIL                & CLAM                 & TransMIL                      & DGRMIL                        & PathPT                        \\ \hline
\multirow{3}{*}{\begin{tabular}[c]{@{}c@{}}PLIP\\ 0.111 (0.100, 0.124)\end{tabular}}  & 1-shot  & 0.103 (0.097, 0.106) & 0.099 (0.086, 0.111) & \textcolor{blue}{0.214 (0.184, 0.236)}    & \textbf{0.217 (0.199, 0.237)} & 0.192 (0.179, 0.196)          \\
                                                                                      & 5-shot  & 0.360 (0.353, 0.368) & 0.358 (0.341, 0.381) & \textcolor{blue}{0.409 (0.398, 0.419)}    & \textbf{0.424 (0.409, 0.433)} & 0.252 (0.245, 0.265)          \\
                                                                                      & 10-shot & 0.419 (0.414, 0.425) & 0.410 (0.403, 0.419) & \textcolor{blue}{0.488 (0.471, 0.496)}    & \textbf{0.491 (0.474, 0.514)} & 0.251 (0.229, 0.276)          \\ \hline
\multirow{3}{*}{\begin{tabular}[c]{@{}c@{}}MUSK\\ 0.253 (0.228, 0.268)\end{tabular}}  & 1-shot  & 0.046 (0.038, 0.053) & 0.094 (0.081, 0.108) & \textbf{0.284 (0.241, 0.309)} & \textcolor{blue}{0.242 (0.227, 0.259)}    & 0.033 (0.033, 0.039)          \\
                                                                                      & 5-shot  & 0.352 (0.332, 0.369) & 0.381 (0.374, 0.391) & \textcolor{blue}{0.493 (0.482, 0.513)}    & \textbf{0.501 (0.492, 0.509)} & 0.410 (0.394, 0.421)          \\
                                                                                      & 10-shot & 0.403 (0.398, 0.429) & 0.442 (0.437, 0.448) & \textcolor{blue}{0.582 (0.578, 0.590)}    & \textbf{0.569 (0.565, 0.584)} & 0.519 (0.512, 0.527)          \\ \hline
\multirow{3}{*}{\begin{tabular}[c]{@{}c@{}}CONCH\\ 0.204 (0.183, 0.224)\end{tabular}} & 1-shot  & 0.097 (0.082, 0.107) & 0.091 (0.081, 0.096) & \textbf{0.343 (0.330, 0.348)} & \textcolor{blue}{0.336 (0.309, 0.347)}    & 0.213 (0.198, 0.220)          \\
                                                                                      & 5-shot  & 0.476 (0.464, 0.488) & 0.477 (0.469, 0.492) & \textbf{0.574 (0.569, 0.586)} & \textcolor{blue}{0.569 (0.552, 0.582)}    & 0.416 (0.406, 0.454)          \\
                                                                                      & 10-shot & 0.542 (0.538, 0.544) & 0.549 (0.543, 0.558) & \textbf{0.621 (0.613, 0.627)} & \textcolor{blue}{0.621 (0.599, 0.637)}    & 0.491 (0.487, 0.499)          \\ \hline
\multirow{3}{*}{\begin{tabular}[c]{@{}c@{}}KEEP\\ 0.408 (0.402, 0.415)\end{tabular}}  & 1-shot  & 0.163 (0.153, 0.164) & 0.199 (0.181, 0.225) & 0.397 (0.387, 0.402)          & \textcolor{blue}{0.416 (0.394, 0.430)}    & \textbf{0.463 (0.445, 0.479)} \\
                                                                                      & 5-shot  & 0.573 (0.563, 0.576) & 0.561 (0.552, 0.574) & 0.587 (0.574, 0.612)          & \textcolor{blue}{0.599 (0.592, 0.611)}    & \textbf{0.629 (0.617, 0.642)} \\
                                                                                      & 10-shot & 0.631 (0.620, 0.642) & 0.629 (0.620, 0.637) & 0.648 (0.637, 0.656)          & \textcolor{blue}{0.650 (0.647, 0.655)}    & \textbf{0.679 (0.666, 0.689)} \\ \hline
\label{tab:ebrains}
\end{tabular}
}
\end{table}

\begin{table}[]
\small
\centering
\caption{\textbf{Experimental results on TCGA-SARC.} Performance comparison of PathPT and various MIL methods across different base models on the TCGA-SARC. Values represent the median of 10 runs, with Q1 and Q3 quartiles shown in parentheses. Bold indicates the best performance and \textcolor{blue}{blue} indicates the second-best performance. MUSK is excluded from this dataset as it is pretrained on TCGA datasets.}
\scalebox{0.75}{
\setlength{\tabcolsep}{{5pt}}
\begin{tabular}{ccccccc}
\hline
                                                                                      &         & ABMIL                & CLAM                 & TransMIL                      & DGRMIL                        & PathPT                        \\ \hline
\multirow{3}{*}{\begin{tabular}[c]{@{}c@{}}PLIP\\ 0.318 (0.301, 0.333)\end{tabular}}  & 1-shot  & 0.257 (0.238, 0.294) & 0.213 (0.209, 0.215) & 0.280 (0.266, 0.317)          & \textcolor{blue}{0.302 (0.290, 0.340)}    & \textbf{0.340 (0.306, 0.355)} \\
                                                                                      & 5-shot  & 0.420 (0.402, 0.467) & 0.453 (0.427, 0.470) & \textbf{0.501 (0.479, 0.528)} & \textcolor{blue}{0.484 (0.466, 0.516)}    & 0.436 (0.374, 0.460)          \\
                                                                                      & 10-shot & 0.478 (0.462, 0.499) & 0.533 (0.494, 0.562) & \textbf{0.583 (0.570, 0.598)} & \textcolor{blue}{0.573 (0.558, 0.577)}    & 0.458 (0.423, 0.496)          \\ \hline
\multirow{3}{*}{\begin{tabular}[c]{@{}c@{}}CONCH\\ 0.257 (0.253, 0.276)\end{tabular}} & 1-shot  & 0.318 (0.285, 0.331) & 0.277 (0.264, 0.291) & \textcolor{blue}{0.356 (0.344, 0.385)}    & \textbf{0.415 (0.397, 0.428)} & 0.286 (0.250, 0.331)          \\
                                                                                      & 5-shot  & 0.528 (0.513, 0.556) & 0.523 (0.506, 0.535) & \textbf{0.580 (0.561, 0.594)} & \textcolor{blue}{0.579 (0.575, 0.594)}    & 0.524 (0.507, 0.577)          \\
                                                                                      & 10-shot & 0.585 (0.543, 0.617) & 0.615 (0.604, 0.630) & 0.615 (0.604, 0.630)          & \textcolor{blue}{0.624 (0.605, 0.630)}    & \textbf{0.641 (0.610, 0.650)} \\ \hline
\multirow{3}{*}{\begin{tabular}[c]{@{}c@{}}KEEP\\ 0.451 (0.438, 0.475)\end{tabular}}  & 1-shot  & 0.200 (0.200, 0.200) & 0.224 (0.215, 0.227) & \textcolor{blue}{0.438 (0.420, 0.465)}    & 0.407 (0.378, 0.432)          & \textbf{0.497 (0.477, 0.519)} \\
                                                                                      & 5-shot  & 0.385 (0.200, 0.446) & 0.509 (0.483, 0.539) & \textcolor{blue}{0.633 (0.597, 0.647)}    & 0.602 (0.583, 0.634)          & \textbf{0.635 (0.591, 0.659)} \\
                                                                                      & 10-shot & 0.581 (0.564, 0.604) & 0.597 (0.579, 0.611) & \textcolor{blue}{0.666 (0.652, 0.685)}    & 0.662 (0.651, 0.669)          & \textbf{0.682 (0.676, 0.688)} \\ \hline
\label{tab:sarc}
\end{tabular}
}
\end{table}

\begin{table}[]
\small
\centering
\caption{\textbf{Experimental results on TCGA-THYM.} Performance comparison of PathPT and various MIL methods across different base models on the TCGA-THYM. Values represent the median of 10 runs, with Q1 and Q3 quartiles shown in parentheses. Bold indicates the best performance and \textcolor{blue}{blue} indicates the second-best performance. MUSK is excluded from this dataset as it is pretrained on TCGA datasets.}
\scalebox{0.75}{
\setlength{\tabcolsep}{{5pt}}
\begin{tabular}{ccccccc}
\hline
                                                                                      &         & ABMIL                & CLAM                 & TransMIL                      & DGRMIL                        & PathPT                        \\ \hline
\multirow{3}{*}{\begin{tabular}[c]{@{}c@{}}PLIP\\ 0.346 (0.299, 0.421)\end{tabular}}  & 1-shot  & 0.250 (0.250, 0.250) & 0.258 (0.244, 0.273) & \textcolor{blue}{0.347 (0.287, 0.376)}    & \textbf{0.363 (0.278, 0.417)} & 0.323 (0.298, 0.362)          \\
                                                                                      & 5-shot  & 0.468 (0.426, 0.504) & 0.511 (0.456, 0.554) & \textcolor{blue}{0.586 (0.560, 0.588)}    & \textbf{0.619 (0.559, 0.626)} & 0.423 (0.364, 0.476)          \\
                                                                                      & 10-shot & 0.625 (0.619, 0.631) & 0.628 (0.614, 0.659) & \textbf{0.683 (0.664, 0.692)} & \textcolor{blue}{0.648 (0.642, 0.689)}    & 0.430 (0.392, 0.459)          \\ \hline
\multirow{3}{*}{\begin{tabular}[c]{@{}c@{}}CONCH\\ 0.290 (0.235, 0.335)\end{tabular}} & 1-shot  & 0.247 (0.213, 0.258) & 0.142 (0.133, 0.156) & 0.278 (0.240, 0.305)          & \textcolor{blue}{0.384 (0.325, 0.472)}    & \textbf{0.433 (0.398, 0.494)} \\
                                                                                      & 5-shot  & 0.555 (0.538, 0.587) & 0.484 (0.465, 0.599) & \textcolor{blue}{0.632 (0.589, 0.662)}    & \textbf{0.664 (0.591, 0.672)} & 0.605 (0.569, 0.698)          \\
                                                                                      & 10-shot & 0.720 (0.700, 0.735) & 0.711 (0.694, 0.721) & \textcolor{blue}{0.733 (0.714, 0.748)}    & 0.715 (0.679, 0.722)          & \textbf{0.777 (0.735, 0.805)} \\ \hline
\multirow{3}{*}{\begin{tabular}[c]{@{}c@{}}KEEP\\ 0.291 (0.239, 0.346)\end{tabular}}  & 1-shot  & 0.256 (0.250, 0.268) & 0.250 (0.227, 0.263) & \textcolor{blue}{0.439 (0.423, 0.459)}    & \textbf{0.441 (0.398, 0.477)} & 0.396 (0.329, 0.416)          \\
                                                                                      & 5-shot  & 0.552 (0.528, 0.623) & 0.561 (0.518, 0.626) & \textbf{0.700 (0.650, 0.728)} & \textcolor{blue}{0.664 (0.640, 0.726)}    & 0.599 (0.534, 0.673)          \\
                                                                                      & 10-shot & 0.691 (0.684, 0.697) & 0.702 (0.681, 0.721) & \textcolor{blue}{0.760 (0.748, 0.777)}    & \textbf{0.763 (0.742, 0.770)} & 0.745 (0.732, 0.762)          \\ \hline
\label{tab:thym}
\end{tabular}
}
\end{table}

\begin{table}[]
\small
\centering
\caption{\textbf{Experimental results on TCGA-UCS.} Performance comparison of PathPT and various MIL methods across different base models on the TCGA-UCS. Values represent the median of 10 runs, with Q1 and Q3 quartiles shown in parentheses. Bold indicates the best performance and \textcolor{blue}{blue} indicates the second-best performance. MUSK is excluded from this dataset as it is pretrained on TCGA datasets.}
\scalebox{0.75}{
\setlength{\tabcolsep}{{5pt}}
\begin{tabular}{ccccccc}
\hline
                                                                                      &         & ABMIL                         & CLAM                          & TransMIL                   & DGRMIL                     & PathPT                        \\ \hline
\multirow{3}{*}{\begin{tabular}[c]{@{}c@{}}PLIP\\ 0.500 (0.500, 0.500)\end{tabular}}  & 1-shot  & \textbf{0.646 (0.576, 0.724)} & 0.458 (0.438, 0.477)          & \textcolor{blue}{0.531 (0.500, 0.555)} & 0.490 (0.419, 0.581)       & 0.510 (0.477, 0.539)          \\
                                                                                      & 5-shot  & \textbf{0.771 (0.706, 0.818)} & 0.615 (0.500, 0.763)          & \textcolor{blue}{0.661 (0.521, 0.766)} & 0.604 (0.490, 0.742)       & 0.583 (0.526, 0.695)          \\
                                                                                      & 10-shot & \textbf{0.771 (0.688, 0.771)} & \textcolor{blue}{0.766 (0.716, 0.771)}    & 0.667 (0.516, 0.727)       & 0.583 (0.565, 0.648)       & 0.583 (0.492, 0.672)          \\ \hline
\multirow{3}{*}{\begin{tabular}[c]{@{}c@{}}CONCH\\ 0.500 (0.500, 0.562)\end{tabular}} & 1-shot  & 0.500 (0.484, 0.547)          & \textcolor{blue}{0.531 (0.490, 0.586)}    & 0.500 (0.500, 0.500)       & 0.521 (0.492, 0.583)       & \textbf{0.562 (0.513, 0.578)} \\
                                                                                      & 5-shot  & \textbf{0.667 (0.583, 0.688)} & 0.625 (0.547, 0.771)          & 0.651 (0.607, 0.680)       & \textcolor{blue}{0.656 (0.602, 0.716)} & 0.651 (0.638, 0.706)          \\
                                                                                      & 10-shot & 0.562 (0.539, 0.578)          & 0.641 (0.586, 0.674)          & \textcolor{blue}{0.677 (0.562, 0.727)} & 0.635 (0.591, 0.690)       & \textbf{0.682 (0.604, 0.729)} \\ \hline
\multirow{3}{*}{\begin{tabular}[c]{@{}c@{}}KEEP\\ 0.510 (0.492, 0.549)\end{tabular}}  & 1-shot  & \textbf{0.656 (0.646, 0.677)} & \textcolor{blue}{0.542 (0.516, 0.594)}    & 0.531 (0.508, 0.539)       & 0.458 (0.354, 0.563)       & 0.510 (0.479, 0.581)          \\
                                                                                      & 5-shot  & 0.604 (0.594, 0.622)          & 0.594 (0.521, 0.680)          & 0.568 (0.534, 0.701)       & \textcolor{blue}{0.615 (0.497, 0.716)} & \textbf{0.651 (0.602, 0.656)} \\
                                                                                      & 10-shot & 0.635 (0.599, 0.661)          & \textbf{0.719 (0.633, 0.737)} & 0.635 (0.602, 0.664)       & 0.646 (0.630, 0.703)       & \textcolor{blue}{0.693 (0.638, 0.758)}    \\ \hline
\label{tab:ucs}
\end{tabular}
}
\end{table}

\begin{table}[]
\small
\centering
\caption{\textbf{Experimental results on Nephroblastoma.} Performance comparison of PathPT and various MIL methods across different base models on the Nephroblastoma. Values represent the median of 10 runs, with Q1 and Q3 quartiles shown in parentheses. Bold indicates the best performance and \textcolor{blue}{blue} indicates the second-best performance.}
\scalebox{0.75}{
\setlength{\tabcolsep}{{5pt}}
\begin{tabular}{ccccccc}
\hline
                                                                                      &         & ABMIL                & CLAM                 & TransMIL                      & DGRMIL                        & PathPT                        \\ \hline
\multirow{3}{*}{\begin{tabular}[c]{@{}c@{}}PLIP\\ 0.235 (0.202, 0.257)\end{tabular}}  & 1-shot  & 0.284 (0.249, 0.294) & 0.256 (0.225, 0.269) & \textcolor{blue}{0.301 (0.257, 0.375)}    & 0.297 (0.260, 0.355)          & \textbf{0.305 (0.299, 0.318)} \\
                                                                                      & 5-shot  & 0.376 (0.358, 0.411) & 0.348 (0.325, 0.381) & \textbf{0.411 (0.400, 0.438)} & \textcolor{blue}{0.391 (0.362, 0.414)}    & 0.302 (0.283, 0.327)          \\
                                                                                      & 10-shot & 0.419 (0.397, 0.430) & 0.427 (0.423, 0.438) & \textbf{0.478 (0.463, 0.500)} & \textcolor{blue}{0.434 (0.396, 0.456)}    & 0.395 (0.333, 0.433)          \\ \hline
\multirow{3}{*}{\begin{tabular}[c]{@{}c@{}}MUSK\\ 0.278 (0.239, 0.291)\end{tabular}}  & 1-shot  & 0.244 (0.208, 0.264) & 0.203 (0.201, 0.203) & \textbf{0.332 (0.260, 0.403)} & \textcolor{blue}{0.306 (0.279, 0.342)}    & 0.200 (0.200, 0.231)          \\
                                                                                      & 5-shot  & 0.310 (0.283, 0.320) & 0.350 (0.330, 0.363) & \textcolor{blue}{0.428 (0.409, 0.465)}    & 0.399 (0.368, 0.442)          & \textbf{0.440 (0.429, 0.492)} \\
                                                                                      & 10-shot & 0.404 (0.385, 0.429) & 0.426 (0.413, 0.446) & \textcolor{blue}{0.493 (0.484, 0.497)}    & 0.493 (0.467, 0.506)          & \textbf{0.508 (0.491, 0.521)} \\ \hline
\multirow{3}{*}{\begin{tabular}[c]{@{}c@{}}CONCH\\ 0.280 (0.243, 0.325)\end{tabular}} & 1-shot  & 0.247 (0.237, 0.260) & 0.200 (0.200, 0.200) & 0.267 (0.232, 0.330)          & \textcolor{blue}{0.271 (0.249, 0.369)}    & \textbf{0.317 (0.308, 0.325)} \\
                                                                                      & 5-shot  & 0.322 (0.315, 0.340) & 0.323 (0.320, 0.332) & \textcolor{blue}{0.422 (0.399, 0.447)}    & 0.403 (0.386, 0.434)          & \textbf{0.476 (0.457, 0.494)} \\
                                                                                      & 10-shot & 0.423 (0.417, 0.436) & 0.421 (0.416, 0.441) & 0.470 (0.455, 0.488)          & \textcolor{blue}{0.484 (0.446, 0.497)}    & \textbf{0.504 (0.485, 0.511)} \\ \hline
\multirow{3}{*}{\begin{tabular}[c]{@{}c@{}}KEEP\\ 0.265 (0.248, 0.271)\end{tabular}}  & 1-shot  & 0.200 (0.200, 0.200) & 0.205 (0.203, 0.212) & \textcolor{blue}{0.291 (0.261, 0.322)}    & \textbf{0.334 (0.272, 0.388)} & 0.264 (0.223, 0.322)          \\
                                                                                      & 5-shot  & 0.341 (0.334, 0.357) & 0.338 (0.318, 0.348) & 0.416 (0.399, 0.437)          & \textcolor{blue}{0.423 (0.408, 0.458)}    & \textbf{0.459 (0.435, 0.480)} \\
                                                                                      & 10-shot & 0.423 (0.399, 0.432) & 0.400 (0.393, 0.415) & 0.471 (0.459, 0.480)          & \textcolor{blue}{0.475 (0.464, 0.481)}    & \textbf{0.491 (0.480, 0.511)} \\ \hline
\label{tab:nephro}
\end{tabular}
}
\end{table}

\begin{table}[]
\small
\centering
\caption{\textbf{Experimental results on Hepatoblastoma.} Performance comparison of PathPT and various MIL methods across different base models on the Hepatoblastoma. Values represent the median of 10 runs, with Q1 and Q3 quartiles shown in parentheses. Bold indicates the best performance and \textcolor{blue}{blue} indicates the second-best performance.}
\scalebox{0.75}{
\setlength{\tabcolsep}{{5pt}}
\begin{tabular}{ccccccc}
\hline
                                                                                      &         & ABMIL                & CLAM                       & TransMIL                      & DGRMIL                        & PathPT                        \\ \hline
\multirow{3}{*}{\begin{tabular}[c]{@{}c@{}}PLIP\\ 0.278 (0.240, 0.294)\end{tabular}}  & 1-shot  & 0.271 (0.266, 0.274) & 0.270 (0.269, 0.275)       & 0.266 (0.236, 0.294)          & \textbf{0.303 (0.270, 0.392)} & \textcolor{blue}{0.281 (0.236, 0.305)}    \\
                                                                                      & 5-shot  & 0.298 (0.293, 0.325) & 0.299 (0.290, 0.321)       & \textcolor{blue}{0.377 (0.349, 0.390)}    & \textbf{0.380 (0.306, 0.430)} & 0.288 (0.258, 0.312)          \\
                                                                                      & 10-shot & 0.370 (0.341, 0.372) & 0.363 (0.351, 0.379)       & \textcolor{blue}{0.384 (0.370, 0.419)}    & \textbf{0.435 (0.398, 0.460)} & 0.349 (0.323, 0.356)          \\ \hline
\multirow{3}{*}{\begin{tabular}[c]{@{}c@{}}MUSK\\ 0.395 (0.315, 0.424)\end{tabular}}  & 1-shot  & 0.254 (0.252, 0.257) & 0.227 (0.222, 0.244)       & \textcolor{blue}{0.278 (0.250, 0.300)}    & \textbf{0.286 (0.261, 0.310)} & 0.250 (0.250, 0.287)          \\
                                                                                      & 5-shot  & 0.265 (0.229, 0.291) & 0.313 (0.296, 0.337)       & \textcolor{blue}{0.401 (0.374, 0.423)}    & 0.327 (0.267, 0.355)          & \textbf{0.428 (0.400, 0.440)} \\
                                                                                      & 10-shot & 0.262 (0.258, 0.269) & 0.378 (0.368, 0.393)       & \textbf{0.447 (0.437, 0.456)} & 0.436 (0.415, 0.450)          & \textcolor{blue}{0.438 (0.423, 0.459)}    \\ \hline
\multirow{3}{*}{\begin{tabular}[c]{@{}c@{}}CONCH\\ 0.276 (0.251, 0.306)\end{tabular}} & 1-shot  & 0.254 (0.246, 0.255) & \textcolor{blue}{0.287 (0.283, 0.293)} & 0.274 (0.235, 0.317)          & 0.282 (0.264, 0.341)          & \textbf{0.424 (0.379, 0.462)} \\
                                                                                      & 5-shot  & 0.308 (0.305, 0.311) & 0.363 (0.331, 0.374)       & \textcolor{blue}{0.437 (0.429, 0.467)}    & 0.391 (0.349, 0.416)          & \textbf{0.515 (0.487, 0.541)} \\
                                                                                      & 10-shot & 0.376 (0.362, 0.382) & 0.374 (0.365, 0.389)       & \textcolor{blue}{0.518 (0.475, 0.534)}    & 0.491 (0.451, 0.527)          & \textbf{0.534 (0.519, 0.560)} \\ \hline
\multirow{3}{*}{\begin{tabular}[c]{@{}c@{}}KEEP\\ 0.313 (0.303, 0.315)\end{tabular}}  & 1-shot  & 0.250 (0.250, 0.250) & 0.218 (0.208, 0.247)       & \textcolor{blue}{0.312 (0.294, 0.316)}    & 0.264 (0.243, 0.296)          & \textbf{0.392 (0.342, 0.425)} \\
                                                                                      & 5-shot  & 0.334 (0.310, 0.344) & 0.297 (0.284, 0.309)       & \textcolor{blue}{0.409 (0.385, 0.422)}    & 0.403 (0.367, 0.420)          & \textbf{0.517 (0.509, 0.549)} \\
                                                                                      & 10-shot & 0.383 (0.377, 0.395) & 0.378 (0.374, 0.392)       & \textcolor{blue}{0.492 (0.466, 0.502)}    & 0.444 (0.434, 0.457)          & \textbf{0.524 (0.502, 0.566)} \\ \hline
\label{tab:hepato}
\end{tabular}
}
\end{table}

\begin{table}[]
\small
\centering
\caption{\textbf{Experimental results on Medulloblastoma.} Performance comparison of PathPT and various MIL methods across different base models on the Medulloblastoma. Values represent the median of 10 runs, with Q1 and Q3 quartiles shown in parentheses. Bold indicates the best performance and \textcolor{blue}{blue} indicates the second-best performance.}
\scalebox{0.75}{
\setlength{\tabcolsep}{{5pt}}
\begin{tabular}{ccccccc}
\hline
                                                                                      &         & ABMIL                         & CLAM                          & TransMIL                      & DGRMIL                     & PathPT                        \\ \hline
\multirow{3}{*}{\begin{tabular}[c]{@{}c@{}}PLIP\\ 0.369 (0.339, 0.393)\end{tabular}}  & 1-shot  & \textbf{0.358 (0.333, 0.397)} & \textcolor{blue}{0.356 (0.353, 0.359)}    & 0.350 (0.266, 0.374)          & 0.327 (0.294, 0.341)       & 0.351 (0.321, 0.378)          \\
                                                                                      & 5-shot  & 0.388 (0.340, 0.442)          & 0.359 (0.331, 0.374)          & \textcolor{blue}{0.427 (0.382, 0.447)}    & 0.404 (0.389, 0.440)       & \textbf{0.513 (0.472, 0.563)} \\
                                                                                      & 10-shot & \textcolor{blue}{0.509 (0.461, 0.601)}    & 0.434 (0.420, 0.464)          & 0.427 (0.398, 0.445)          & 0.403 (0.385, 0.436)       & \textbf{0.592 (0.524, 0.617)} \\ \hline
\multirow{3}{*}{\begin{tabular}[c]{@{}c@{}}MUSK\\ 0.398 (0.334, 0.426)\end{tabular}}  & 1-shot  & 0.333 (0.333, 0.333)          & \textbf{0.514 (0.501, 0.542)} & \textcolor{blue}{0.450 (0.426, 0.474)}    & 0.416 (0.361, 0.440)       & 0.333 (0.333, 0.406)          \\
                                                                                      & 5-shot  & 0.333 (0.333, 0.333)          & 0.572 (0.513, 0.602)          & \textbf{0.623 (0.577, 0.661)} & \textcolor{blue}{0.591 (0.551, 0.646)} & 0.536 (0.519, 0.597)          \\
                                                                                      & 10-shot & 0.465 (0.449, 0.497)          & 0.614 (0.588, 0.629)          & \textcolor{blue}{0.658 (0.627, 0.688)}    & 0.650 (0.629, 0.674)       & \textbf{0.659 (0.643, 0.690)} \\ \hline
\multirow{3}{*}{\begin{tabular}[c]{@{}c@{}}CONCH\\ 0.325 (0.309, 0.364)\end{tabular}} & 1-shot  & 0.367 (0.350, 0.380)          & \textcolor{blue}{0.426 (0.387, 0.441)}    & 0.383 (0.363, 0.413)          & 0.425 (0.326, 0.471)       & \textbf{0.515 (0.386, 0.548)} \\
                                                                                      & 5-shot  & 0.388 (0.370, 0.441)          & 0.500 (0.456, 0.551)          & \textbf{0.595 (0.574, 0.605)} & 0.553 (0.535, 0.565)       & \textcolor{blue}{0.592 (0.582, 0.675)}    \\
                                                                                      & 10-shot & 0.528 (0.471, 0.554)          & 0.577 (0.556, 0.591)          & \textcolor{blue}{0.579 (0.533, 0.632)}    & 0.572 (0.535, 0.601)       & \textbf{0.581 (0.544, 0.608)} \\ \hline
\multirow{3}{*}{\begin{tabular}[c]{@{}c@{}}KEEP\\ 0.380 (0.369, 0.400)\end{tabular}}  & 1-shot  & \textcolor{blue}{0.465 (0.393, 0.518)}    & 0.332 (0.297, 0.352)          & 0.397 (0.326, 0.451)          & 0.444 (0.404, 0.543)       & \textbf{0.494 (0.390, 0.528)} \\
                                                                                      & 5-shot  & 0.506 (0.485, 0.573)          & 0.496 (0.482, 0.540)          & \textcolor{blue}{0.568 (0.557, 0.607)}    & 0.547 (0.539, 0.578)       & \textbf{0.627 (0.595, 0.637)} \\
                                                                                      & 10-shot & \textbf{0.593 (0.544, 0.612)} & 0.572 (0.559, 0.589)          & \textcolor{blue}{0.576 (0.556, 0.611)}    & 0.542 (0.504, 0.579)       & 0.564 (0.517, 0.621)          \\ \hline
\label{tab:medullo}
\end{tabular}
}
\end{table}

\begin{table}[]
\small
\centering
\caption{\textbf{Experimental results on Neuroblastoma.} Performance comparison of PathPT and various MIL methods across different base models on the Neuroblastoma. Values represent the median of 10 runs, with Q1 and Q3 quartiles shown in parentheses. Bold indicates the best performance and \textcolor{blue}{blue} indicates the second-best performance.}
\scalebox{0.75}{
\setlength{\tabcolsep}{{5pt}}
\begin{tabular}{ccccccc}
\hline
                                                                                      &         & ABMIL                & CLAM                 & TransMIL                      & DGRMIL                        & PathPT                        \\ \hline
\multirow{3}{*}{\begin{tabular}[c]{@{}c@{}}PLIP\\ 0.395 (0.336, 0.440)\end{tabular}}  & 1-shot  & 0.335 (0.322, 0.388) & 0.369 (0.357, 0.377) & \textcolor{blue}{0.399 (0.367, 0.423)}    & \textbf{0.451 (0.383, 0.465)} & 0.350 (0.336, 0.428)          \\
                                                                                      & 5-shot  & 0.481 (0.425, 0.510) & 0.407 (0.360, 0.459) & \textbf{0.536 (0.470, 0.545)} & \textcolor{blue}{0.530 (0.509, 0.603)}    & 0.474 (0.461, 0.512)          \\
                                                                                      & 10-shot & 0.530 (0.503, 0.550) & 0.496 (0.466, 0.527) & \textbf{0.578 (0.564, 0.610)} & \textcolor{blue}{0.578 (0.554, 0.611)}    & 0.501 (0.462, 0.512)          \\ \hline
\multirow{3}{*}{\begin{tabular}[c]{@{}c@{}}MUSK\\ 0.427 (0.400, 0.490)\end{tabular}}  & 1-shot  & 0.333 (0.333, 0.333) & 0.381 (0.374, 0.385) & \textcolor{blue}{0.417 (0.406, 0.426)}    & \textbf{0.467 (0.445, 0.508)} & 0.404 (0.341, 0.442)          \\
                                                                                      & 5-shot  & 0.333 (0.333, 0.333) & 0.514 (0.452, 0.526) & \textcolor{blue}{0.540 (0.508, 0.598)}    & \textbf{0.577 (0.498, 0.594)} & 0.531 (0.518, 0.559)          \\
                                                                                      & 10-shot & 0.554 (0.333, 0.577) & 0.513 (0.464, 0.529) & \textbf{0.605 (0.596, 0.620)} & \textcolor{blue}{0.585 (0.540, 0.592)}    & 0.554 (0.534, 0.566)          \\ \hline
\multirow{3}{*}{\begin{tabular}[c]{@{}c@{}}CONCH\\ 0.353 (0.337, 0.450)\end{tabular}} & 1-shot  & 0.414 (0.392, 0.429) & 0.393 (0.381, 0.400) & \textcolor{blue}{0.474 (0.445, 0.556)}    & 0.458 (0.420, 0.475)          & \textbf{0.514 (0.468, 0.519)} \\
                                                                                      & 5-shot  & 0.468 (0.427, 0.487) & 0.505 (0.379, 0.513) & \textbf{0.564 (0.511, 0.570)} & 0.518 (0.478, 0.548)          & \textcolor{blue}{0.520 (0.481, 0.557)}    \\
                                                                                      & 10-shot & 0.431 (0.409, 0.464) & 0.496 (0.453, 0.534) & \textcolor{blue}{0.572 (0.535, 0.586)}    & \textbf{0.578 (0.535, 0.582)} & 0.493 (0.466, 0.560)          \\ \hline
\multirow{3}{*}{\begin{tabular}[c]{@{}c@{}}KEEP\\ 0.551 (0.535, 0.555)\end{tabular}}  & 1-shot  & 0.418 (0.400, 0.429) & 0.333 (0.333, 0.333) & \textcolor{blue}{0.453 (0.424, 0.461)}    & 0.423 (0.402, 0.438)          & \textbf{0.490 (0.478, 0.518)} \\
                                                                                      & 5-shot  & 0.468 (0.439, 0.522) & 0.336 (0.317, 0.446) & \textcolor{blue}{0.527 (0.492, 0.570)}    & 0.517 (0.502, 0.543)          & \textbf{0.568 (0.555, 0.592)} \\
                                                                                      & 10-shot & 0.513 (0.493, 0.532) & 0.497 (0.472, 0.549) & \textcolor{blue}{0.566 (0.526, 0.577)}    & 0.541 (0.502, 0.572)          & \textbf{0.583 (0.569, 0.591)} \\ \hline
\label{tab:neuro}
\end{tabular}
}
\end{table}

\begin{table}[]
\small
\centering
\caption{\textbf{Experimental results on UBC-OCEAN.} Performance comparison of PathPT and various MIL methods across different base models on the UBC-OCEAN. Values represent the median of 10 runs, with Q1 and Q3 quartiles shown in parentheses. Bold indicates the best performance and \textcolor{blue}{blue} indicates the second-best performance.}
\scalebox{0.75}{
\setlength{\tabcolsep}{{5pt}}
\begin{tabular}{ccccccc}
\hline
                                                                                      &         & ABMIL                & CLAM                 & TransMIL                      & DGRMIL                        & PathPT                        \\ \hline
\multirow{3}{*}{\begin{tabular}[c]{@{}c@{}}PLIP\\ 0.320 (0.265, 0.380)\end{tabular}}  & 1-shot  & 0.190 (0.182, 0.198) & 0.190 (0.183, 0.207) & \textcolor{blue}{0.310 (0.265, 0.337)}    & \textbf{0.310 (0.280, 0.337)} & 0.245 (0.205, 0.305)          \\
                                                                                      & 5-shot  & 0.430 (0.410, 0.467) & 0.440 (0.410, 0.472) & \textcolor{blue}{0.520 (0.512, 0.570)}    & \textbf{0.525 (0.493, 0.565)} & 0.435 (0.430, 0.455)          \\
                                                                                      & 10-shot & 0.565 (0.522, 0.608) & 0.570 (0.520, 0.588) & \textbf{0.645 (0.605, 0.670)} & \textcolor{blue}{0.630 (0.575, 0.657)}    & 0.510 (0.485, 0.520)          \\ \hline
\multirow{3}{*}{\begin{tabular}[c]{@{}c@{}}MUSK\\ 0.520 (0.417, 0.562)\end{tabular}}  & 1-shot  & 0.200 (0.200, 0.200) & 0.200 (0.200, 0.200) & \textbf{0.390 (0.337, 0.407)} & \textcolor{blue}{0.315 (0.280, 0.387)}    & 0.200 (0.200, 0.232)          \\
                                                                                      & 5-shot  & 0.365 (0.342, 0.387) & 0.435 (0.408, 0.487) & \textcolor{blue}{0.640 (0.608, 0.665)}    & 0.575 (0.545, 0.588)          & \textbf{0.675 (0.605, 0.690)} \\
                                                                                      & 10-shot & 0.570 (0.537, 0.587) & 0.610 (0.572, 0.618) & \textcolor{blue}{0.720 (0.687, 0.730)}    & 0.700 (0.570, 0.708)          & \textbf{0.730 (0.722, 0.748)} \\ \hline
\multirow{3}{*}{\begin{tabular}[c]{@{}c@{}}CONCH\\ 0.375 (0.352, 0.407)\end{tabular}} & 1-shot  & 0.270 (0.262, 0.290) & 0.220 (0.213, 0.220) & 0.360 (0.325, 0.388)          & \textcolor{blue}{0.465 (0.433, 0.507)}    & \textbf{0.560 (0.520, 0.627)} \\
                                                                                      & 5-shot  & 0.485 (0.480, 0.522) & 0.520 (0.490, 0.545) & 0.665 (0.632, 0.687)          & \textcolor{blue}{0.670 (0.652, 0.670)}    & \textbf{0.740 (0.730, 0.768)} \\
                                                                                      & 10-shot & 0.590 (0.560, 0.618) & 0.605 (0.592, 0.618) & 0.710 (0.700, 0.745)          & \textcolor{blue}{0.715 (0.703, 0.728)}    & \textbf{0.790 (0.790, 0.808)} \\ \hline
\multirow{3}{*}{\begin{tabular}[c]{@{}c@{}}KEEP\\ 0.660 (0.650, 0.667)\end{tabular}}  & 1-shot  & 0.200 (0.200, 0.200) & 0.300 (0.273, 0.310) & 0.445 (0.395, 0.510)          & \textcolor{blue}{0.445 (0.423, 0.487)}    & \textbf{0.510 (0.443, 0.613)} \\
                                                                                      & 5-shot  & 0.610 (0.562, 0.630) & 0.545 (0.525, 0.600) & \textcolor{blue}{0.730 (0.680, 0.763)}    & 0.710 (0.625, 0.725)          & \textbf{0.795 (0.775, 0.810)} \\
                                                                                      & 10-shot & 0.755 (0.750, 0.767) & 0.730 (0.705, 0.755) & \textcolor{blue}{0.795 (0.783, 0.807)}    & 0.795 (0.772, 0.808)          & \textbf{0.820 (0.802, 0.837)} \\ \hline
\label{tab:ubc}
\end{tabular}
}
\end{table}

\begin{table}[]
\small
\centering
\caption{\textbf{Experimental results on TCGA-BRCA.} Performance comparison of PathPT and various MIL methods across different base models on the TCGA-BRCA dataset. Values represent the median of 10 runs, with Q1 and Q3 quartiles shown in parentheses. \textbf{Bold} indicates the best performance and \textcolor{blue}{blue} indicates the second-best performance. MUSK is excluded from this dataset as it is pretrained on TCGA datasets.}
\scalebox{0.75}{
\setlength{\tabcolsep}{{5pt}}
\begin{tabular}{ccccccc}
\hline
                                                                                      &         & ABMIL                      & CLAM                       & TransMIL                      & DGRMIL                        & PathPT                        \\ \hline
\multirow{3}{*}{\begin{tabular}[c]{@{}c@{}}PLIP\\ 0.546 (0.510, 0.550)\end{tabular}}  & 1-shot  & 0.479 (0.475, 0.498)       & 0.479 (0.475, 0.490)       & 0.500 (0.500, 0.500)          & \textbf{0.575 (0.552, 0.596)} & \textcolor{blue}{0.533 (0.512, 0.569)}    \\
                                                                                      & 5-shot  & 0.487 (0.463, 0.521)       & 0.554 (0.521, 0.575)       & \textcolor{blue}{0.579 (0.492, 0.633)}    & \textbf{0.588 (0.521, 0.631)} & 0.575 (0.546, 0.598)          \\
                                                                                      & 10-shot & 0.567 (0.496, 0.660)       & \textcolor{blue}{0.646 (0.625, 0.658)} & \textbf{0.650 (0.627, 0.658)} & 0.621 (0.590, 0.665)          & 0.554 (0.529, 0.594)          \\ \hline
\multirow{3}{*}{\begin{tabular}[c]{@{}c@{}}CONCH\\ 0.596 (0.544, 0.706)\end{tabular}} & 1-shot  & 0.500 (0.485, 0.506)       & \textcolor{blue}{0.500 (0.500, 0.500)} & 0.487 (0.467, 0.512)          & 0.408 (0.371, 0.460)          & \textbf{0.700 (0.621, 0.715)} \\
                                                                                      & 5-shot  & 0.558 (0.521, 0.617)       & 0.500 (0.500, 0.500)       & 0.692 (0.619, 0.760)          & \textcolor{blue}{0.704 (0.608, 0.750)}    & \textbf{0.783 (0.733, 0.804)} \\
                                                                                      & 10-shot & 0.692 (0.683, 0.710)       & 0.721 (0.700, 0.742)       & \textcolor{blue}{0.771 (0.760, 0.790)}    & 0.762 (0.752, 0.767)          & \textbf{0.787 (0.777, 0.813)} \\ \hline
\multirow{3}{*}{\begin{tabular}[c]{@{}c@{}}KEEP\\ 0.767 (0.758, 0.767)\end{tabular}}  & 1-shot  & \textcolor{blue}{0.575 (0.560, 0.608)} & 0.550 (0.500, 0.565)       & 0.492 (0.444, 0.500)          & 0.517 (0.500, 0.606)          & \textbf{0.708 (0.606, 0.773)} \\
                                                                                      & 5-shot  & 0.642 (0.606, 0.690)       & 0.646 (0.600, 0.692)       & 0.700 (0.671, 0.742)          & \textcolor{blue}{0.713 (0.671, 0.742)}    & \textbf{0.812 (0.802, 0.833)} \\
                                                                                      & 10-shot & 0.721 (0.710, 0.748)       & 0.729 (0.725, 0.756)       & \textcolor{blue}{0.767 (0.758, 0.781)}    & 0.750 (0.750, 0.758)          & \textbf{0.842 (0.833, 0.860)} \\ \hline
\label{tab:brca}
\end{tabular}
}
\end{table}

\begin{table}[]
\small
\centering
\caption{\textbf{Experimental results on TCGA-BRAIN.} Performance comparison of PathPT and various MIL methods across different base models on the TCGA-BRAIN. Values represent the median of 10 runs, with Q1 and Q3 quartiles shown in parentheses. \textbf{Bold} indicates the best performance and \textcolor{blue}{blue} indicates the second-best performance. MUSK is excluded from this dataset as it is pretrained on TCGA datasets.}
\scalebox{0.75}{
\setlength{\tabcolsep}{{5pt}}
\begin{tabular}{ccccccc}
\hline
                                                                                      &         & ABMIL                & CLAM                 & TransMIL                   & DGRMIL                        & PathPT                        \\ \hline
\multirow{3}{*}{\begin{tabular}[c]{@{}c@{}}PLIP\\ 0.394 (0.379, 0.417)\end{tabular}}  & 1-shot  & 0.333 (0.304, 0.337) & 0.333 (0.328, 0.344) & 0.311 (0.292, 0.321)       & \textcolor{blue}{0.381 (0.322, 0.400)}    & \textbf{0.383 (0.368, 0.426)} \\
                                                                                      & 5-shot  & 0.414 (0.371, 0.435) & 0.372 (0.311, 0.425) & 0.450 (0.419, 0.475)       & \textcolor{blue}{0.456 (0.421, 0.471)}    & \textbf{0.497 (0.479, 0.511)} \\
                                                                                      & 10-shot & 0.547 (0.511, 0.560) & 0.528 (0.513, 0.532) & 0.547 (0.529, 0.575)       & \textbf{0.597 (0.556, 0.615)} & \textcolor{blue}{0.553 (0.535, 0.565)}    \\ \hline
\multirow{3}{*}{\begin{tabular}[c]{@{}c@{}}CONCH\\ 0.411 (0.374, 0.460)\end{tabular}} & 1-shot  & 0.353 (0.346, 0.361) & 0.383 (0.379, 0.392) & 0.339 (0.324, 0.353)       & \textbf{0.450 (0.410, 0.478)} & \textcolor{blue}{0.428 (0.414, 0.442)}    \\
                                                                                      & 5-shot  & 0.450 (0.432, 0.476) & 0.481 (0.451, 0.510) & \textcolor{blue}{0.594 (0.558, 0.610)} & 0.553 (0.533, 0.582)          & \textbf{0.628 (0.622, 0.658)} \\
                                                                                      & 10-shot & 0.572 (0.568, 0.600) & 0.561 (0.544, 0.571) & \textcolor{blue}{0.683 (0.674, 0.694)} & 0.672 (0.661, 0.697)          & \textbf{0.694 (0.689, 0.704)} \\ \hline
\multirow{3}{*}{\begin{tabular}[c]{@{}c@{}}KEEP\\ 0.603 (0.594, 0.606)\end{tabular}}  & 1-shot  & 0.356 (0.350, 0.399) & 0.347 (0.335, 0.356) & 0.378 (0.343, 0.400)       & \textcolor{blue}{0.406 (0.386, 0.426)}    & \textbf{0.475 (0.444, 0.512)} \\
                                                                                      & 5-shot  & 0.475 (0.457, 0.489) & 0.494 (0.482, 0.518) & 0.594 (0.572, 0.617)       & \textcolor{blue}{0.611 (0.578, 0.633)}    & \textbf{0.653 (0.635, 0.672)} \\
                                                                                      & 10-shot & 0.581 (0.561, 0.601) & 0.631 (0.612, 0.650) & \textcolor{blue}{0.681 (0.667, 0.694)} & 0.678 (0.662, 0.688)          & \textbf{0.697 (0.689, 0.729)} \\ \hline
\label{tab:brain}
\end{tabular}
}
\end{table}

\begin{table}[]
\small
\centering
\caption{\textbf{Experimental results of tumor segmentation.} Dice performance of PathPT across different datasets and different base models on the tumor segmentation task. Results are presented using the same format with Table \ref{tab:brca}.}
\scalebox{0.9}{
\setlength{\tabcolsep}{{5pt}}
\begin{tabular}{ccccc}
\hline
                       &         & CAMELYON16                    & PANDA                         & AGGC22                        \\ \hline
\multirow{4}{*}{PLIP}  & 0-shot  & 0.141 (0.106, 0.154)          & 0.255 (0.165, 0.337)          & 0.166 (0.060, 0.218)          \\
                       & 1-shot  & 0.175 (0.168, 0.264)          & \textbf{0.456 (0.359, 0.463)} & 0.433 (0.401, 0.461)          \\
                       & 5-shot  & \textcolor{blue}{0.359 (0.344, 0.394)}    & 0.419 (0.397, 0.447)          & \textcolor{blue}{0.497 (0.483, 0.515)}    \\
                       & 10-shot & \textbf{0.384 (0.373, 0.394)} & \textcolor{blue}{0.444 (0.420, 0.463)}    & \textbf{0.530 (0.521, 0.536)} \\ \hline
\multirow{4}{*}{MUSK}  & 0-shot  & 0.136 (0.130, 0.152)          & 0.269 (0.159, 0.345)          & 0.242 (0.151, 0.300)          \\
                       & 1-shot  & 0.245 (0.127, 0.392)          & 0.362 (0.240, 0.442)          & 0.505 (0.493, 0.540)          \\
                       & 5-shot  & \textbf{0.504 (0.427, 0.569)} & \textcolor{blue}{0.405 (0.362, 0.444)}    & \textcolor{blue}{0.578 (0.567, 0.585)}    \\
                       & 10-shot & \textcolor{blue}{0.459 (0.428, 0.492)}    & \textbf{0.455 (0.436, 0.473)} & \textbf{0.592 (0.585, 0.597)} \\ \hline
\multirow{4}{*}{CONCH} & 0-shot  & 0.129 (0.123, 0.149)          & 0.302 (0.228, 0.324)          & 0.239 (0.208, 0.280)          \\
                       & 1-shot  & 0.421 (0.271, 0.497)          & 0.435 (0.376, 0.450)          & 0.570 (0.563, 0.598)          \\
                       & 5-shot  & \textbf{0.522 (0.491, 0.550)} & \textcolor{blue}{0.439 (0.423, 0.448)}    & \textcolor{blue}{0.618 (0.611, 0.626)}    \\
                       & 10-shot & \textcolor{blue}{0.505 (0.489, 0.514)}    & \textbf{0.466 (0.461, 0.487)} & \textbf{0.624 (0.623, 0.633)} \\ \hline
\multirow{4}{*}{KEEP}  & 0-shot  & 0.207 (0.134, 0.231)          & 0.327 (0.143, 0.369)          & 0.272 (0.178, 0.341)          \\
                       & 1-shot  & 0.377 (0.337, 0.475)          & 0.419 (0.351, 0.454)          & 0.604 (0.591, 0.632)          \\
                       & 5-shot  & \textbf{0.591 (0.558, 0.615)} & \textcolor{blue}{0.460 (0.426, 0.484)}    & \textcolor{blue}{0.634 (0.630, 0.641)}    \\
                       & 10-shot & \textcolor{blue}{0.560 (0.547, 0.574)}    & \textbf{0.515 (0.479, 0.534)} & \textbf{0.629 (0.627, 0.636)} \\ \hline
\label{tab:seg}
\end{tabular}
}
\end{table}

\begin{table}[]
\small
\centering
\caption{\textbf{Ablation study results of tumor segmentation.} Ablation dice performance of PathPT, CoOp (without spatial-aware module) and visual feature linear probing (without prompt learning) across different datasets and different base models on tumor segmentation task. Results are presented using the same format with Table \ref{tab:brca}.}
\scalebox{0.9}{
\setlength{\tabcolsep}{{5pt}}
\begin{tabular}{cccccc}
\hline
                            &                                                                                       &         & Linear Probing             & CoOp                       & PathPT                        \\ \hline
\multirow{6}{*}{CAMELYON16} & \multirow{3}{*}{\begin{tabular}[c]{@{}c@{}}CONCH\\ 0.129 (0.123, 0.149)\end{tabular}} & 1-shot  & \textcolor{blue}{0.118 (0.030, 0.284)} & 0.048 (0.048, 0.049)       & \textbf{0.421 (0.271, 0.497)} \\
                            &                                                                                       & 5-shot  & \textcolor{blue}{0.384 (0.292, 0.468)} & 0.153 (0.152, 0.163)       & \textbf{0.522 (0.491, 0.550)} \\
                            &                                                                                       & 10-shot & \textcolor{blue}{0.350 (0.332, 0.392)} & 0.217 (0.216, 0.228)       & \textbf{0.505 (0.489, 0.514)} \\ \cline{2-6} 
                            & \multirow{3}{*}{\begin{tabular}[c]{@{}c@{}}KEEP\\ 0.207 (0.134, 0.231)\end{tabular}}  & 1-shot  & 0.193 (0.164, 0.297)       & \textcolor{blue}{0.225 (0.225, 0.226)} & \textbf{0.377 (0.337, 0.475)} \\
                            &                                                                                       & 5-shot  & \textcolor{blue}{0.492 (0.393, 0.502)} & 0.268 (0.258, 0.274)       & \textbf{0.591 (0.558, 0.615)} \\
                            &                                                                                       & 10-shot & \textcolor{blue}{0.407 (0.396, 0.446)} & 0.301 (0.285, 0.319)       & \textbf{0.560 (0.547, 0.574)} \\ \hline
\multirow{6}{*}{PANDA}      & \multirow{3}{*}{\begin{tabular}[c]{@{}c@{}}CONCH\\ 0.302 (0.228, 0.324)\end{tabular}} & 1-shot  & \textcolor{blue}{0.280 (0.000, 0.397)} & 0.049 (0.039, 0.058)       & \textbf{0.435 (0.376, 0.450)} \\
                            &                                                                                       & 5-shot  & 0.004 (0.000, 0.304)       & \textcolor{blue}{0.006 (0.002, 0.007)} & \textbf{0.439 (0.423, 0.448)} \\
                            &                                                                                       & 10-shot & \textcolor{blue}{0.257 (0.133, 0.329)} & 0.006 (0.005, 0.007)       & \textbf{0.466 (0.461, 0.487)} \\ \cline{2-6} 
                            & \multirow{3}{*}{\begin{tabular}[c]{@{}c@{}}KEEP\\ 0.327 (0.143, 0.369)\end{tabular}}  & 1-shot  & 0.312 (0.002, 0.399)       & \textcolor{blue}{0.337 (0.333, 0.341)} & \textbf{0.419 (0.351, 0.454)} \\
                            &                                                                                       & 5-shot  & 0.322 (0.228, 0.404)       & \textcolor{blue}{0.341 (0.318, 0.344)} & \textbf{0.460 (0.426, 0.484)} \\
                            &                                                                                       & 10-shot & \textcolor{blue}{0.421 (0.405, 0.425)} & 0.355 (0.349, 0.361)       & \textbf{0.515 (0.479, 0.534)} \\ \hline
\multirow{6}{*}{AGGC22}     & \multirow{3}{*}{\begin{tabular}[c]{@{}c@{}}CONCH\\ 0.239 (0.208, 0.280)\end{tabular}} & 1-shot  & \textcolor{blue}{0.227 (0.057, 0.419)} & 0.019 (0.018, 0.021)       & \textbf{0.570 (0.563, 0.598)} \\
                            &                                                                                       & 5-shot  & \textcolor{blue}{0.468 (0.432, 0.525)} & 0.105 (0.075, 0.132)       & \textbf{0.618 (0.611, 0.626)} \\
                            &                                                                                       & 10-shot & \textcolor{blue}{0.480 (0.460, 0.520)} & 0.444 (0.439, 0.446)       & \textbf{0.624 (0.623, 0.633)} \\ \cline{2-6} 
                            & \multirow{3}{*}{\begin{tabular}[c]{@{}c@{}}KEEP\\ 0.272 (0.178, 0.341)\end{tabular}}  & 1-shot  & \textcolor{blue}{0.354 (0.232, 0.551)} & 0.310 (0.308, 0.310)       & \textbf{0.604 (0.591, 0.632)} \\
                            &                                                                                       & 5-shot  & \textcolor{blue}{0.598 (0.557, 0.615)} & 0.428 (0.417, 0.440)       & \textbf{0.634 (0.630, 0.641)} \\
                            &                                                                                       & 10-shot & \textcolor{blue}{0.599 (0.596, 0.608)} & 0.501 (0.498, 0.510)       & \textbf{0.629 (0.627, 0.636)} \\ \hline
\label{tab:seg_abl}
\end{tabular}
}
\end{table}

\begin{table}[]
\small
\centering
\caption{Dataset statistics of adult rare tumor from EBRAINS.}
\scalebox{1.}{
\setlength{\tabcolsep}{{5pt}}
\begin{tabular}{clcccc}
\hline
Dataset                   & \multicolumn{1}{c}{Subtype}                                   & Train Set & Test Set & \multicolumn{2}{c}{Total}  \\ \hline
\multirow{30}{*}{EBRAINS} & Ependymoma                                                    & 15        & 15       & 30 & \multirow{30}{*}{898} \\
                          & Medulloblastoma, non-WNT/non-SHH                              & 15        & 15       & 30 &                       \\
                          & Pituitary adenoma                                             & 15        & 15       & 30 &                       \\
                          & Gliosarcoma                                                   & 15        & 15       & 30 &                       \\
                          & Anaplastic astrocytoma, IDH-mutant                            & 15        & 15       & 30 &                       \\
                          & Meningothelial meningioma                                     & 14        & 15       & 29 &                       \\
                          & Diffuse astrocytoma, IDH-mutant                               & 15        & 15       & 30 &                       \\
                          & Transitional meningioma                                       & 15        & 15       & 30 &                       \\
                          & Diffuse large B-cell lymphoma of the CNS                      & 15        & 15       & 30 &                       \\
                          & Lipoma                                                        & 15        & 15       & 30 &                       \\
                          & Metastatic tumours                                            & 15        & 15       & 30 &                       \\
                          & Fibrous meningioma                                            & 15        & 15       & 30 &                       \\
                          & Ganglioglioma                                                 & 15        & 15       & 30 &                       \\
                          & Glioblastoma, IDH-wildtype                                    & 15        & 15       & 30 &                       \\
                          & Haemangioma                                                   & 15        & 15       & 30 &                       \\
                          & Atypical meningioma                                           & 15        & 15       & 30 &                       \\
                          & Secretory meningioma                                          & 15        & 15       & 30 &                       \\
                          & Anaplastic astrocytoma, IDH-wildtype                          & 15        & 15       & 30 &                       \\
                          & Angiomatous meningioma                                        & 15        & 15       & 30 &                       \\
                          & Haemangioblastoma                                             & 15        & 15       & 30 &                       \\
                          & Anaplastic ependymoma                                         & 15        & 15       & 30 &                       \\
                          & Adamantinomatous craniopharyngioma                            & 15        & 15       & 30 &                       \\
                          & Anaplastic meningioma                                         & 15        & 15       & 30 &                       \\
                          & Anaplastic oligodendroglioma, IDH-mutant and 1p/19q codeleted & 15        & 15       & 30 &                       \\
                          & Oligodendroglioma, IDH-mutant and 1p/19q codeleted            & 15        & 15       & 30 &                       \\
                          & Haemangiopericytoma                                           & 15        & 15       & 30 &                       \\
                          & Langerhans cell histiocytosis                                 & 14        & 15       & 29 &                       \\
                          & Schwannoma                                                    & 15        & 15       & 30 &                       \\
                          & Glioblastoma, IDH-mutant                                      & 15        & 15       & 30 &                       \\
                          & Pilocytic astrocytoma                                         & 15        & 15       & 30 &                       \\ \hline
\label{tab:data_ebrains}
\end{tabular}
}
\end{table}

\begin{table}[]
\small
\centering
\caption{Dataset statistics of adult rare tumor from TCGA.}
\scalebox{1.}{
\setlength{\tabcolsep}{{5pt}}
\begin{tabular}{clcccc}
\hline
Dataset                    & \multicolumn{1}{c}{Subtype}              & Train Set & Test Set & \multicolumn{2}{c}{Total}  \\ \hline
\multirow{5}{*}{TCGA-SARC} & Malignant Peripheral Nerve Sheath Tumors & 15        & 12       & 27  & \multirow{5}{*}{577} \\
                           & Dedifferentiated liposarcoma             & 15        & 73       & 88  &                      \\
                           & Myxofibrosarcoma                         & 15        & 126      & 141 &                      \\
                           & Undifferentiated Pleomorphic Sarcoma     & 15        & 151      & 166 &                      \\
                           & Leiomyosarcoma                           & 15        & 140      & 155 &                      \\ \hline
\multirow{4}{*}{TCGA-THYM} & Type B1                                  & 15        & 21       & 36  & \multirow{4}{*}{152} \\
                           & Type A                                   & 15        & 10       & 25  &                      \\
                           & Type B2                                  & 15        & 30       & 45  &                      \\
                           & Type AB                                  & 15        & 31       & 46  &                      \\ \hline
\multirow{2}{*}{TCGA-UCS}  & Heterologous Type                        & 15        & 16       & 31  & \multirow{2}{*}{52}  \\
                           & Homologous Type                          & 15        & 6        & 21  &                      \\ \hline
\label{tab:data_adult}
\end{tabular}
}
\end{table}

\begin{table}[]
\small
\centering
\caption{Dataset statistics of pediatric rare cancer.}
\scalebox{1.}{
\setlength{\tabcolsep}{{5pt}}
\begin{tabular}{clcccc}
\hline
Dataset                          & \multicolumn{1}{c}{Subtype}                            & Train Set & Test Set & \multicolumn{2}{c}{Total}  \\ \hline
\multirow{3}{*}{Neuroblastoma}   & Ganglioneuroblastoma, intermixed                       & 15        & 17       & 32  & \multirow{3}{*}{136} \\
                                 & Poorly differentiated neuroblastoma                    & 15        & 54       & 69  &                      \\
                                 & Differentiating neuroblastoma                          & 15        & 20       & 35  &                      \\ \hline
\multirow{3}{*}{Medulloblastoma} & Classic medulloblastoma                                & 15        & 182      & 197 & \multirow{3}{*}{238} \\
                                 & Large Cell/Anaplastic medulloblastoma                  & 5         & 6        & 11  &                      \\
                                 & Desmoplastic nodular medulloblastoma                   & 15        & 15       & 30  &                      \\ \hline
\multirow{4}{*}{Hepatoblastoma}  & Epithelial macrotrabecular pattern of hepatoblastoma   & 5         & 5        & 10  & \multirow{4}{*}{439} \\
                                 & Pure fetal hepatoblastoma with low mitotic activity    & 14        & 164      & 178 &                      \\
                                 & Mixed epithelial and mesenchymal hepatoblastoma        & 15        & 61       & 76  &                      \\
                                 & Epithelial mixed fetal and embryonal hepatoblastoma    & 15        & 160      & 175 &                      \\ \hline
\multirow{5}{*}{Nephroblastoma}  & Mixed blastemal, epithelial and stromal nephroblastoma & 15        & 189      & 204 & \multirow{5}{*}{416} \\
                                 & Mixed blastemal and stromal nephroblastoma             & 15        & 38       & 53  &                      \\
                                 & Mixed blastemal and epithelial nephroblastoma          & 15        & 42       & 57  &                      \\
                                 & Nephroblastoma, stromal type                           & 15        & 75       & 90  &                      \\
                                 & Nephroblastoma, blastemal type                         & 5         & 7        & 12  &                      \\ \hline
\label{tab:data_child}
\end{tabular}
}
\end{table}

\begin{table}[]
\small
\centering
\caption{Dataset statistics of common cancer.}
\scalebox{1.}{
\setlength{\tabcolsep}{{5pt}}
\begin{tabular}{clcccc}
\hline
Dataset                     & \multicolumn{1}{c}{Subtype} & Train Set & Test Set & \multicolumn{2}{c}{Total} \\ \hline
\multirow{2}{*}{TCGA-BRCA}  & Invasive Ductal Carcinoma   & 15        & 60       & 75 & \multirow{2}{*}{150} \\
                            & Invasive Lobular Carcinoma  & 15        & 60       & 75 &                      \\ \hline
\multirow{3}{*}{TCGA-BRAIN} & Glioblastoma                & 15        & 60       & 75 & \multirow{3}{*}{223} \\
                            & Oligodendroglioma           & 13        & 60       & 73 &                      \\
                            & Astrocytoma                 & 15        & 60       & 75 &                      \\ \hline
\multirow{5}{*}{UBC-OCEAN}  & EC                          & 15        & 20       & 35 & \multirow{5}{*}{175} \\
                            & HGSC                        & 15        & 20       & 35 &                      \\
                            & CC                          & 15        & 20       & 35 &                      \\
                            & MC                          & 15        & 20       & 35 &                      \\
                            & LGSC                        & 15        & 20       & 35 &                      \\ \hline
\label{tab:data_common}
\end{tabular}
}
\end{table}

\begin{table}[]
\small
\centering
\caption{Dataset statistics of tumor segmentation.}
\scalebox{1.}{
\setlength{\tabcolsep}{{5pt}}
\begin{tabular}{cccc}
\hline
Dataset    & Train Set & Test Set & Total \\ \hline
Camelyon16 & 15        & 33       & 48    \\ \hline
AGGC22     & 15        & 113      & 128   \\ \hline
PANDA      & 15        & 10580    & 10595 \\ \hline
\label{tab:data_seg}
\end{tabular}
}
\end{table}






\end{document}